\documentclass[square,numbers]{article}

\usepackage[preprint]{neurips_2023}

\usepackage[utf8]{inputenc} 
\usepackage[T1]{fontenc}    
\usepackage{hyperref}       
\usepackage{url}            
\usepackage{booktabs}       
\usepackage{amsfonts}       
\usepackage{nicefrac}       
\usepackage{microtype}      
\usepackage{xcolor}         
\usepackage{times}
\usepackage{soul}
\usepackage{graphicx}
\usepackage{amsmath}
\usepackage{amsthm}
\usepackage{amssymb}
\usepackage{algorithm}
\usepackage{algorithmic}
\usepackage{subfigure}
\usepackage{multirow}
\usepackage{wrapfig}
\newtheorem{example}{Example}

\newtheorem{definition}{Definition}
\newtheorem{hypothesis}{Hypotheses}

\title{Towards Causal Representation Learning and Deconfounding from Indefinite Data}

\author{%
  Hang Chen\\
  Department of Computer Science\\
  Xi'an Jiaotong University\\
  China\\
  \texttt{albert2123@stu.xjtu.edu.cn} \\
  \And
  Xinyu Yang \\
  Department of Computer Science\\
  Xi'an Jiaotong University\\
  China\\
  \texttt{yxyphd@mail.xjtu.edu.cn} \\
  \And
  Qing Yang\\
  Department of Data intelligence\\
  Du Xiaoman Financial\\
  China\\
  \texttt{yangqing@duxiaoman.com} \\
}

\begin{document}

\maketitle

\begin{abstract}
  Owing to the cross-pollination between causal discovery 
  and deep learning, non-statistical data (e.g., images, text, etc.) 
  encounters significant conflicts in terms of properties 
  and methods with traditional causal data. To unify 
  these data types of varying forms, 
  we redefine causal data from two novel perspectives and then 
  propose three data paradigms. 
    Among them, the \textit{indefinite data} (like dialogues or 
    video sources) induce low sample utilization and 
    incapability of the distribution assumption, 
    both leading to the fact that learning causal representation 
    from indefinite data is, as of yet, largely unexplored. 
    We design the causal strength variational model to 
    settle down these two problems. Specifically, 
    we leverage the causal strength instead of independent noise 
    as the latent variable to construct evidence lower bound. 
    By this design ethos, The causal strengths of different structures  
    are regarded as a distribution and can be expressed as 
    a 2D matrix. Moreover, considering the 
    latent confounders, we disentangle the causal graph 
    $\mathcal{G}$ into two relation subgraphs 
    $\mathcal{O}$ and $\mathcal{C}$. $\mathcal{O}$ 
    contains pure relations between observed variables, while 
    $\mathcal{C}$ represents the relations from latent variables to 
    observed variables. We implement the above designs as a 
    dynamic variational inference model, 
    tailored to learn causal representation from 
    indefinite data under latent confounding. Finally, we conduct 
    comprehensive experiments on synthetic and real-world data to 
    demonstrate the effectiveness of our method. 
\end{abstract}

\section{Introduction}

Identifying causal relationships from observed data, 
known as Causal Discovery, has drawn much attention in 
the fields leveraging substantial statistical datasets, 
such as the CPRD~\cite{herrett2015data}, 
the DWD Climate~\cite{dietrich2019temporal} and 
AutoMPG~\cite{asuncion2007uci}. However, in light of the 
recent advances in deep learning, 
particularly in LLMs, there is a growing tendency 
to incorporate causal discovery in more complex forms of data, 
including images~\cite{fan2022debiasing}, text~\cite{chen2023affective}, and videos~\cite{oh2021causal}.

These data, sourced from various modalities and non-statistical, 
have lacked a unified definition due to their intricate 
characteristics and complex structures~\cite{scholkopf2021toward}. 
For instance, the Netsim dataset~\cite{smith2011network} introduced 50 distinct 
causal structures. When conventional methods  
~\cite{verma1992algorithm,spirtes2000causation,chickering2002optimal}
are extended to the Netsim dataset, 
they have to refit a new probabilistic model to account for 
a set of new samples whenever the causal structure does not hold. 
Another example is the IEMOCAP dataset~\cite{busso2008iemocap}, 
wherein the causal variables are text type and 
cannot be directly represented numerically. 
In computations, these are often quantified using 
multi-dimensional deep representation (word embeddings). 
Unlike traditional datasets containing statistical data 
such as body temperature, blood pressure, wind strength, etc., 
capitalizing on conventional methods like 
conditional independence~\cite{spirtes2000causation}, 
regression residual~\cite{xie2020generalized}, 
or rank of covariance~\cite{squires2022causal},  
prove challenging for such deep representation  
due to their dimension values' interdependencies.

Hence, trying to give an unified redefinition of 
both emerging datasets and the traditional ones, 
we, from a new perspective, utilize the skeleton quantity $M$ to represent 
the number of involved causal sturctures and adopt the dimensionality 
$D$ of the causal representation to gauge the complexity 
of the one causal variable. The distinction between 
multi-skeleton ($M>1$) and single-skeleton ($M=1$) 
lies in whether multiple causal structures exist within a dataset. 
Simultaneously, the difference between multi-value ($D>1$) and 
single-value ($D=1$) depends on whether the causal variables 
are deep representations or not. 

Consequently, we categorize causal data 
into three data paradigms systematically: 
\textbf{Definite data} characterized by a single-skeleton 
structure ($M=1$) and single-value variables ($D=1$); 
\textbf{Semi-definite data} characterized either by a 
single-skeleton structure ($M=1$) and multi-value variables ($D>1$), 
or by multi-skeleton structures ($M>1$) and single-value variables 
($D=1$); and \textbf{Indefinite data} characterized by 
multi-skeleton structures ($M>1$) and multi-value variables ($D>1$). 

On the extreme end, a number of methods have been proposed to address these two 
challenges, but all of them face diverse limitations. For example, 
to ameliorate the problem of multi-skeleton structures, some methods
~\cite{dhir2020integrating,huang2020causala,lowe2022amortized} 
account for different structures by learning the dynamics across all 
samples, which can infer causal relations in previously unseen samples 
without refitting the model. However, these methods impose 
a stringent assumption of a stationary number of observed nodes which 
is often violated in many applications. Moreover, to model 
multi-value nodes, existing methods
\cite{chen2023affective,fan2022debiasing,wu2022discovering} 
seek to learn a causal representation via auto regression model 
while ignoring the effect of latent confounders. 

Towards causal representation learning and deconfounding from  
multi-skeleton and multi-value data, 
We designed a novel dynamic variational framework. Specifically,

i) We designed a variational model with the causal strength matrix 
as the latent variable. Comparing with the conventional method 
which adopts noise matrix as the latent variable, it has significant 
superiority in processing multi-skeleton and multi-value data. 

ii) We adopted the individual-specific effects on each observed node 
instead of estimation on latent variables to probe 
latent confounding. Under this observe-node-oriented representation, 
we can disentangle the causal graph $\mathcal{G}$ into two subgraphs 
$\mathcal{O}$ and $\mathcal{C}$. 
$\mathcal{O}$ includes the pure relations between observed nodes, 
and $\mathcal{C}$ includes the relations from latent variables 
to observed nodes. We can thus capture the aggressive property 
of confounders and predict the disentangled causal representation. 

Overall, this work contributes 
a redefinition of causal data, 
a design of causal strength probabilistic method, 
a theoretical insight of confounders disentanglement, 
an implementation of the variational model, 
and comprehensive experimental evaluations. 

\section{Definition, Hypothesis, and Background} 

We begin with establishing consistent terminology: 
the terms ``causal model'', ``causal structure'', and ``causal graph'' 
are equivalent, and the prototype of these concepts is 
``causal skeleton.'' (e.g., the causal graph is a DAG, but the 
causal skeleton is a PDAG or others) Similarly, the terms 
``causal variable'' and ``node'' in a causal graph are synonymous, 
as are ``causal representation'' and ``node attribute'', 
both of which represent the value of the causal variable (or node). 
Moreover, each sample within a dataset should depict a causal model 
and therefore possess a corresponding causal skeleton, 
consisting of causal variables. Each causal variable exists 
as a node, and directed edges between nodes represent 
causal relationships. Please note that throughout this paper, 
we assume that causal graph are DAGs, and we do not consider 
the case of maximum ancestor graphs (MAGs). 

\begin{definition}[Causal Data]\footnote{
    The skeleton and variable dimension we propose 
    serve to broaden perspectives on causal data, 
    hence introducing certain conflicts with traditional cognition of 
    causal model. This caused previous reviewers to 
    struggle with conceiving what indefinite data looks like, 
    what constitutes a multi-skeleton structure, 
    and what multi-value variables are. Therefore, 
    we dedicatedly established Appendix A, 
    which elucidates these questions through abundant 
    data examples and details the essential differences 
    among the three data paradigms.}
    The causal relationships exist in a dataset 
    $\mathbf{D} = \{X_{s}\}^{S}_{s=1}$ which has 
    $S$ samples and $M\in \mathbb{N}$ ($M\geqslant 1$) causal structures 
    ($\mathcal{G}=\{\mathcal{E}_{m}, \mathcal{V}_{m}\}^{M}_{m=1}$). 
    Each structure $\mathcal{G}_{m}$ corresponds to several samples separately
    Hence, each sample $X_{s,m} \in \mathbb{R}^{N \times D}$ 
    belongs to a causal structure  
    $\mathcal{G}_{m}=\{\mathcal{E}_{m}, \mathcal{V}_{m}\}$ and 
    consists of $N_{m}$ variables: 
    $X_{s}=\{x_{s,m,n}\}^{N_{m}}_{n_{m}=1}$. 
    $\hat{x}_{s,m,n} \in \mathbb{R}^{1 \times D} (D \geqslant 1)$ 
    represents the causal representation of a varaible $x_{s,m,n}$ 
    where $D$ denotes the dimension of the causal representation. We 
    assume that the number of causal skeletons is equal to the number 
    of causal structures. 
  
    Based on the above datasets, we define three data paradigms: 
  \begin{itemize}
  \item \textbf{Definite Data}: The causal structure is single-skeleton ($M=1$) 
  and the causal variable is single-value ($D=1$).
   
  \item \textbf{Semi-Definite Data}: The causal structure is 
  single-skeleton ($M=1$) and the causal variable is 
  multi-value ($D>1$), or the causal structure is 
  multi-skeleton ($M>1$) and the causal variable is 
  single-value ($D=1$).
    
  \item \textbf{Indefinite Data}: The causal structure is 
  multi-skeleton ($M>1$) and the causal variable is 
  multi-value ($D>1$).
  \end{itemize}

    \label{def2}
  \end{definition}

Moreover, in view of the exsiting prevalent indefinite datasets, 
which mostly include resources like textual conversations 
and video sources, we propose a hypothesis compatible 
with indefinite data: 

\begin{hypothesis}
    The variables of indefinite data satisfy a 
    natural partial order $\preccurlyeq _{X_{s,m}}$, 
    where the explanation of this partial order is, 
    if $x_{1} \preccurlyeq _{X_{s,m}} x_{2}$, 
    it indicates there is no causal relationship  $x_{2} \rightarrow x_{1}$.
    \label{hyp1}
    \end{hypothesis}
    
In most instances, the poset property of Hypothesis~\ref{hyp1}  
satisfies time series, and it is a crucial basis 
for identifying the direction of causal relationships 
when the noise distribution is unknown. 
Causal variables in the poset $\preccurlyeq _{X_{s,m}}$ 
only exist in causal relationships where previous variables 
point to later ones. Of course, such a poset possesses 
transitivity, anti-symmetry, and reflexivity. 
Based on the properties of the poset, 
if we assume non-existence of $x$ loop-itself, 
then this poset correlates directly with DAGs. 
Thus, the majority of indefinite data studies do not need to 
postulate Acyclic restrictions 
(such as the NOTEARS~\cite{zheng2018dags}, etc.), 
and only need to restrict the adjacency matrix 
to be a strictly triangular matrix. 

\begin{figure*}
    \centering
    \includegraphics[width=1\linewidth]{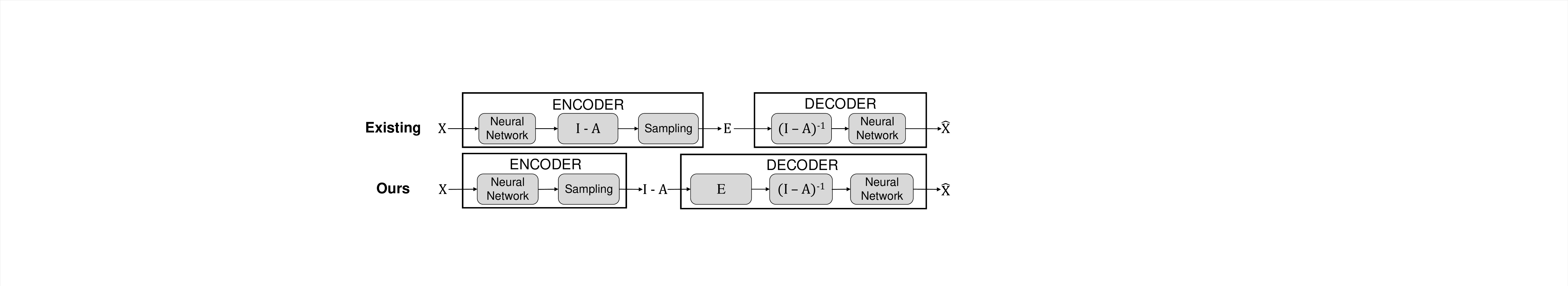}
    \caption{Contrast of two variational frameworks. Both frameworks 
    are constructed by the same autoregression SEMs (Equation 5 
    and 6 in Appendix B). However, our framework adopts $I-A$ as latent 
    variables instead of $E$, in which $A$ represents the causal strength.}
  \label{figtwomodel}
  \end{figure*}

  \subsection{Examples for Three Data Paradigms}

  \begin{example}[Definite Data]
    Arrhythmia Dataset~\cite{647926} is a case record dataset from 
    patients with arrhythmias, including 452 samples, and each sample 
    consists of 279 single-value representations 
    (e.g., age, weight, heart rate, etc.). All samples contribute a 
    common causal graph with 279 nodes, where the edge value indicates 
    some causal relationship, such as the causal strength of how age 
    affects heart rate. 
    \label{exp1}
  \end{example}
  
  \begin{example}[Semi-definite Data (Multi-skeleton and Single-value)]
    The Netsim dataset~\cite{smith2011network} is a simulated fMRI dataset. 
    Because different activities in brain regions over time imply 
    different categories, a set of records of one patient corresponds 
    to one causal skeleton. This dataset includes 50 skeletons and 
    each skeleton consists of 15 nodes that measure the signal 
    strength of 15 brain regions. 
    \label{exp2}
  \end{example} 
  
  \begin{example}[Semi-definite Data (Single-skeleton and Multi-value)]
    CMNIST-75sp~\cite{fan2022debiasing} is a graph classification dataset 
    with controllable bias degrees. In this dataset, all researchers 
    concentrate on one causal graph including 4 variables: causal 
    variable $C$, background variable $B$, observed graph $G$ and label 
    $Y$. $C$ is a part of the MNIST image including multi value of a 
    group of pixels. 
    \label{exp3}
  \end{example}

  \begin{example}[Indefinite Data]
    IEMOCAP Dataset~\cite{busso2008iemocap} is a conversation record dataset 
    with each sample including a dialogue between two speakers. All 100 
    samples are assigned into 26 graphs (i.e., 26 skeletons) based on the 
    speaker identifies and turns and each sample consists of 5-24 
    nodes where the attribute of each node is an 
    utterance represented by $\mathbb{R}^{1 \times 768}$ or 
    $\mathbb{R}^{1 \times 1024}$ in prevalent pretrained language models. 
    \label{exp4}
  \end{example} 
  
  \begin{figure}
    \centering
    \subfigure[definite datasets]{
      \includegraphics[width=0.48\textwidth]{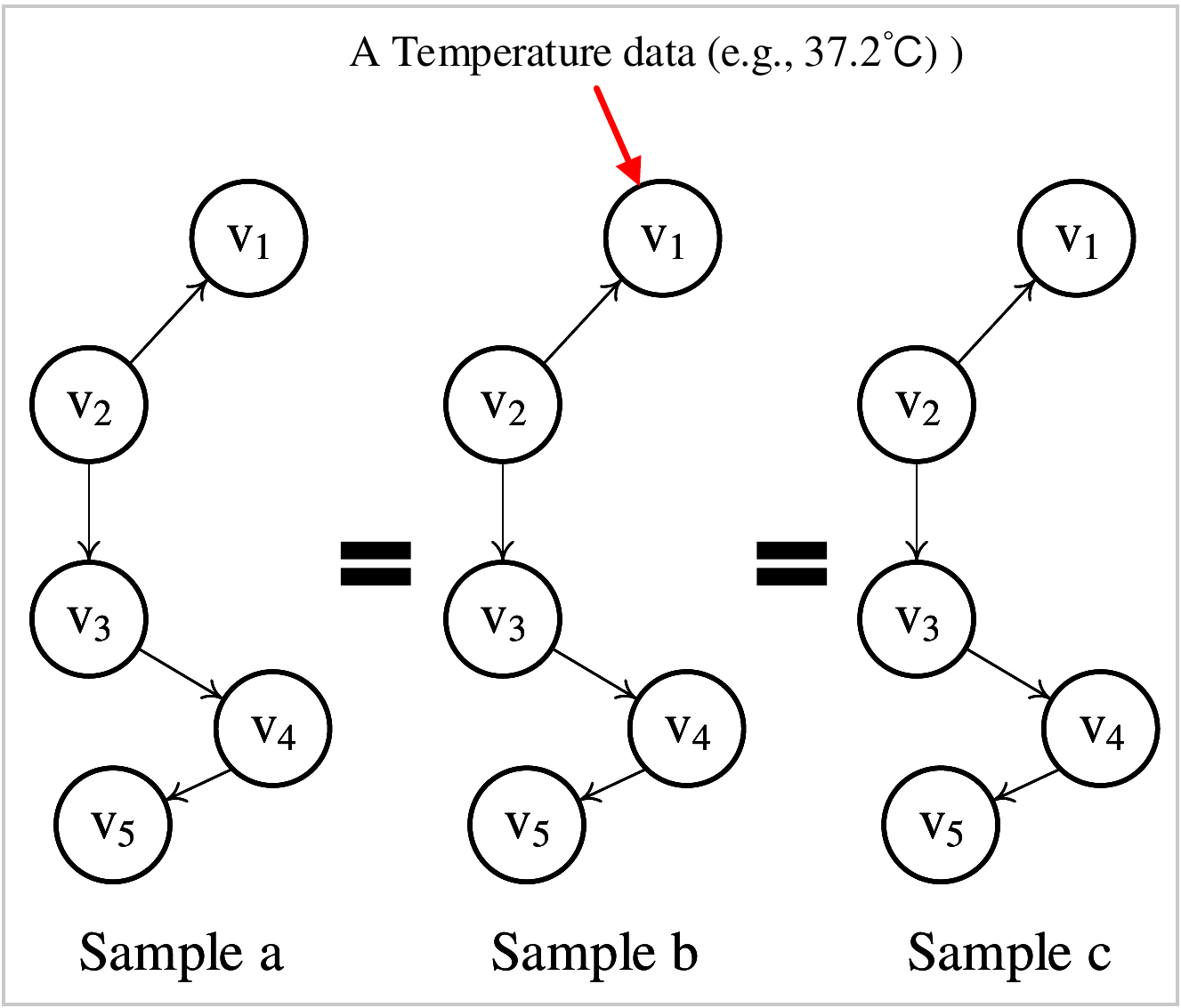}}
    \subfigure[indefinite datasets]{
      \includegraphics[width=0.48\textwidth]{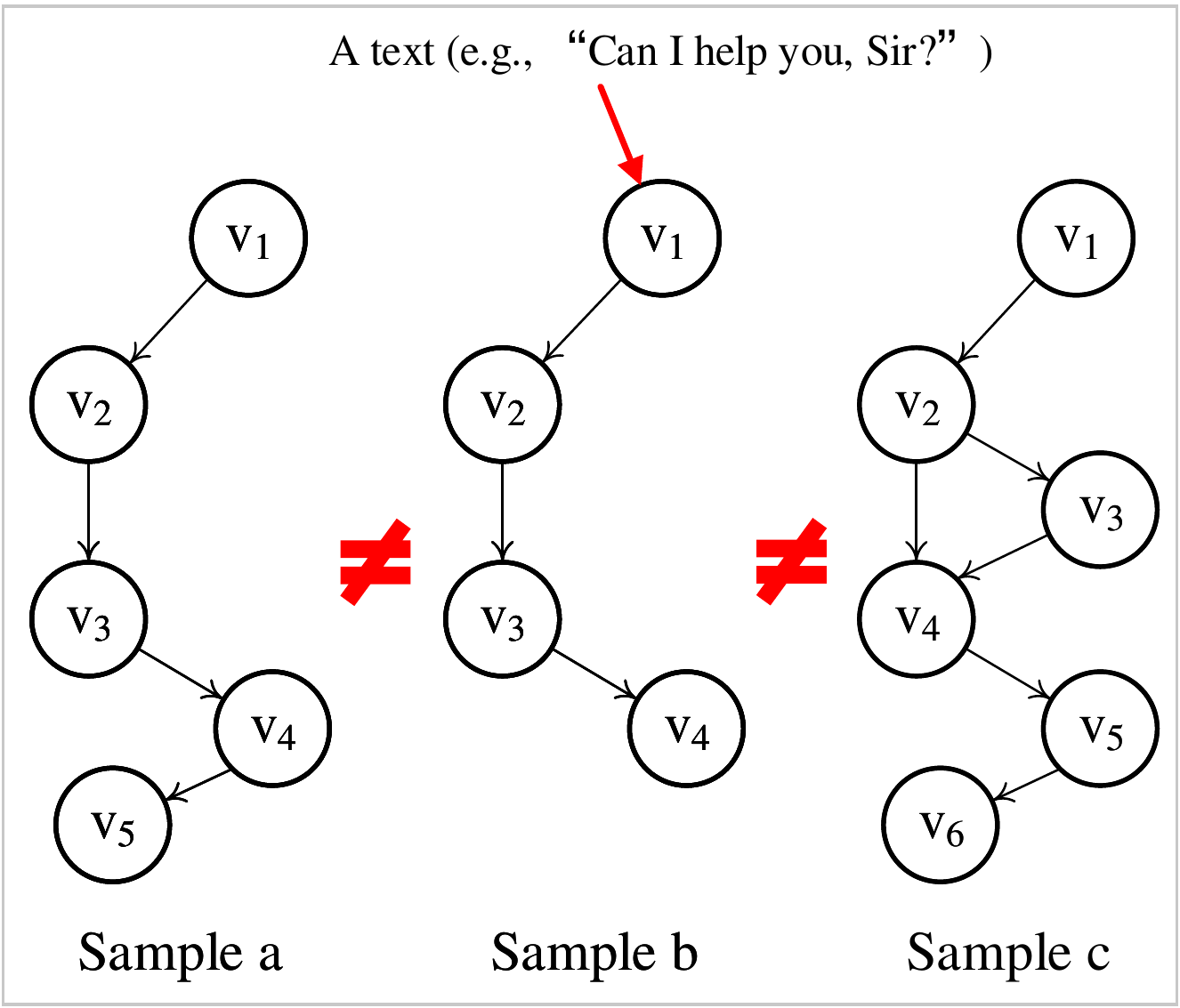}}
    
    \caption{Differences between definite datasets 
    (where $M=1$ and $D=1$) and indefinite datasets 
    (where $M>1$ and $D>1$). In definite datasets, 
    each sample corresponds to an identical causal structure, 
    implying a single-skeleton trait as the entire dataset 
    involves only a single causal structure. In contrast, 
    indefinite datasets do not possess one causal structure 
    for all samples. For instance, there might be varying numbers 
    of causal variables in samples a, b, and c; 
    the relationship $v_{2}\rightarrow v_{4}$ may be absent 
    in sample b but present in sample c. Furthermore, 
    the causal variables in definite and indefinite datasets 
    also differ. For example, in definite datasets, 
    the causal variable $v_{1}$ might represent body temperature, 
    with a causal representation of $37.3$ in sample a, $37.1$ in 
    sample b and $36.8$ in sample c, while $v_{2}$ 
    might symbolize blood pressure, with a causal representation of 
    $118$, $127$, and $135$ separately; they are both single-value data. However, in indefinite 
    data, within a dialogue dataset, the causal variable $v_{1}$ 
    might be an utterance (``Can I help you, Sir?'' in sample a, 
    ``Nice to meet you !'' in sample b, and ``What's the matter with you'' 
    in sample c) 
    with its causal representation being a $768$-dimension 
    word embedding in deep model, and $v_{2}$ might be a 
    responsing utterance to $v_{1}$. In a video dataset, 
    $v_{1}$ might denote a segment representing a particular action 
    or event, with its causal representation as the corresponding 
    optical flow, and $v_{2}$ might be another segment triggered by 
    $v_{1}$.}
    \label{figcomparing}
  \end{figure}

  We aim to illustrate the relationships among three data paradigms 
  through Examples ~\ref{exp1},\ref{exp2},\ref{exp3},\ref{exp4} and Figure~\ref{figcomparing}, 
  focusing particularly on the number of skeletons (single or multi-skeleton) 
  and the dimension of causal representations (single or multi-value). 
  
  \textbf{single or multi-skeleton}: Compared to single-skeleton data, 
  multi-skeleton data lacks discrimination about which samples belong 
  to the same causal model. Therefore, it requires algorithms 
  capable of distinguishing between different causal structures 
  or clustering similar samples. Simultaneously, multi-skeleton data 
  often have trouble in low sample utilization 
  since samples from other skeletons contribute nothing when 
  identifying a specific causal structure. 
  Consequently, the pathways that algorithms focus on single-skeleton 
  and multi-skeleton data are different. 
  
  \textbf{single or multi-value}: multi-value data often facilitate 
  the quantification by deep models, such as text, image, audio, 
  and video, as exemplified in our Figure~\ref{figcomparing}. 
  Compared to single-value data, it involves more complex environments. 
  The statistical strengths of single-value data are more significant, 
  such as computing independence between two single-value 
  representations. On the contrary, determining such ``independence'' 
  among multi-value representations is challenging, 
  often approximated through algorithms like cosine similarity. 
  In Structural Causal Models (SCMs), one can assume that the noise 
  of single-value data follows a specific distribution, 
  but in multi-value data, the noise items are multi-value 
  and interdependent among dimensions, causing many traditional 
  causal discovery methods to make no efforts with multi-value data. 
  
  \subsection{Why We Definite Three Data Paradigms?}
  
  Specifically, we employ the theory illustrated in
  ~\cite{scholkopf2021toward} to explicate why the skeleton and 
  variable dimension are pivotal in engendering differences 
  in causal discovery algorithms. In accordance with the assumption 
  in~\cite{scholkopf2021toward}, the domain of causal variables 
  $\mathcal{X}$ acquires a causal representation $\hat{\mathcal{X}}$ 
  via the encoder $p_{\varphi}$ and decoder $q_{\theta}$, 
  showcasing the causal mechanism in structural equations: 
  
  \begin{equation}
    \hat{x}_{i}=f_{i}(Pa_{i},U_{i})
  \end{equation}
  
  where $Pa_{i}$ represent the parent node set of $x_{i}$. 
  For instance, $p_{\varphi}:U=(1-F)X $ and 
  $q_{\theta}: \hat{X}=(1-F)^{-1}U$. Without prior knowledge, 
  there exist two pathways to recover the causal model: 
  1) Given a fixed causal structure and known causal representation, 
  the causal strength can be estimated by the statistical strength 
  observable in the samples. 
  2) If encoder and decoder are feasible, optimal solutions 
  of the causal model can be achieved by minimizing the 
  reconstruction loss $p_{\varphi}\circ f\circ q_{\theta}$. 
  Here we would like to delimit the solvability of this process 
  in contexts of single-skeleton ($M=1$) and multi-skeleton ($M>1$).
  
  \textbf{For a single-skeleton model}: 
  When the causal structure is fixed, causal strengths $f$ 
  can be calculated. If the causal representation is single-value, 
  the causal structure can be determined without the 
  encoder $p_{\varphi}$ or decoder $q_{\theta}$. 
  The reconstruction loss in this case is $f$. However, 
  for multi-value data, in the reconstruction loss function 
  $p_{\varphi}\circ f\circ q_{\theta}$, $f$ represents the 
  determined part. 
  
  \textbf{For a multi-skeleton model}: 
  The multi-skeleton data induce uncertainty in causal structures, 
  unclear of which samples correspond to the same causal structure 
  and therefore making causal strengths $f$ unsolvable directly. 
  However, under single-value condition with precise quantification 
  processes, the precision of clustering is guaranteed. 
  We can approach by first clustering the samples, 
  and then separate the problem to several tasks of 
  definite data problem-solving ($M=1$, $D=1$). 
  In this regard, reconstruction loss amounts to $\{f\}$, 
  representing the set containing each sub-task's $f$, or 
  in other words, reconstruction loss can be regarded as a 
  multi-task optimization problem, 
  $\alpha 1 f1 + \alpha2 f2 + \alpha3 f3$, where $\alpha$ 
  is the weights of the sample quantity per structure. 
  The worst-case scenario arises with multi-value data, 
  only able to attain an approximate encoder 
  $\hat{p}_{\varphi}=p_{\varphi}\circ f$, 
  which results in a final reconstruction loss of 
  $\hat{p}_{\varphi} \circ q_{\theta}$. Causal strength $f$ 
  comprises an unspecified part. 
  
  In summary, for definite data ($M=1$, $D=1$), 
  it suffices to identify the causal strength 
  between any two causal variables under a certain causal structure. 
  Semi-definite data addresses the problem of discriminating 
  multi-skeleton structures and quantifying multi-value variables 
  separately. As for indefinite data, in the absence of additional 
  assumptions, causal discovery in such datasets presents an 
  ill-posed problem, given it requires deliberation of 
  variable encoding in resolving structure discernibility. 
  
\subsection{Related Works}
Although indefinite data is a new definition proposed by this paper, 
a number of methods have been proposed to address the emerging challenge 
of recovering causal relations from multi-skeleton or multi-value datasets. 
These methods fall into two general categories. First, some methods try to learn an 
invariance for \textit{multi-skeleton data}, such as ACD
~\cite{lowe2022amortized}, which leverages the shared dynamics 
information to train a single model inferring causal relations across 
samples with diverse underlying causal graphs. This range of approaches 
to multi-skeleton data~\cite{yu2019dag,lowe2022amortized,huang2020causala} 
uses neural networks to automate the 
learning of causal structure across examples, limited in the setting 
where plenty of related datasets are cohesive
~\cite{dhir2020integrating,huang2020causala,huang2020causalb,huang2019specific}.  
A second method for \textit{multi-value data} also exists. 
Within the stationary structure of 
discovering causal models on post-partitioned observed nodes, these 
methods try to learn the favorable causal representation of the 
rationalization task. As the most prevailing formulation, these methods 
involve GNN interpretability and transferability
~\cite{fan2022debiasing,wu2022discovering}, conversation relationship 
analysis~\cite{chen2023affective,zhao2022knowledge}, and domain generalization
~\cite{lv2022causality,jiang2022invariant}. But the fundamental equation 
structure does not work when the latent confounders are considered.

Our method inherits the task requirements of these methods, 
aiming to learn causal representations and identify causal models 
to address some high-level causality-related tasks. 
Simultaneously, our approach consider the both multi-skeleton and 
multi-value challenges, as well as the deconfounding 
of latent variables. 

\section{Proposed Probabilistic Framework}\label{AppC}
Many variational models for causal discovery, including
linear SEM variational model~\cite{yu2019dag}, autoregressive
~\cite{wang2020causal} and recently substitude of noise~\cite{chen2023affective}, 
can be encapsulated by a probabilistic framework: 
\begin{itemize}
  \item [1.]
  Construct a Linear Structural Equation Model (SEM) to displace SCM. 
  Specifically, let A $\in \mathbb{R}^{N \times N}$ be the adjacency 
  matrix, and $N$ stands for the number of variables 
  (i.e., the number of attributes from the dataset view 
  or the number of nodes from the graph view). 
  $X\in \mathbb{R}^{N \times 1}$ is a sample of $N$ sample. 
  SCM describes the generation process of a single variable, but it 
  can be trivially represented by a matrix form, where each row of 
  $A$ represents the causal strength from all variables to one 
  observed variable. It is a strong inductive bias that $A$ is a 
  strictly lower triangular due to the acyclic causal order
  ~\cite{shimizu2006linear}.  The linear SEM model reads: 
  \begin{equation} 
    X=AX+E
    \label{eqt3}
   \end{equation}
   where $E\in \mathbb{R}^{N \times 1}$ is the matrix of independent 
   noise $\epsilon_{x_{n}}$. Under this matrix form, generating a 
   random independent noise $E$ can be equivalent to ancestral 
   sampling from the DAG: $X=(I-A)^{-1}E$. 
  \item [2.]
  Build a pair of Autoregression SEMs: 
  \begin{equation} 
    E=(I-A)X
    \label{eqt4}
   \end{equation}
   \begin{equation} 
    X=(I-A)^{-1}E
    \label{eqt5}
   \end{equation}
   Equation~\ref{eqt5} describes a general form as a decoder 
   of a generation model that takes noise $E$ as input and returns $X$ 
   as results and Equation~\ref{eqt4} describes the corresponding 
   encoder. Despite the diverse implementation of parameterized 
   deep learning networks, we without loss of generality adopt $
   f(\cdot)$ to encapsulate the neural network architecture: 
   \begin{equation} 
    E=f_{2}((I-A)f_{1}(X))
    \label{eqt6}
   \end{equation}
   \begin{equation} 
    X=f_{4}((I-A)^{-1}f_{3}(E))
    \label{eqt7}
   \end{equation}
   where $f(\cdot)$ perform nonlinear transforms on $X$ and $E$. 
   Graph neural network (GNN) and Multilayer perceptron (MLP) are 
   popular amplification of function $f(\cdot)$. 
   \item[3.] 
   Considering a specification of noise ($E$) distribution sampling 
   $\{X_{s}\}^{S}_{s=1}$ in definite data, Equation~\ref{eqt7} can 
   be written by a maximization of log-evidence: 
   \begin{equation}
   \frac{1}{S} \sum_{s=1}^{S} \log p(X_{s})=\frac{1}{S} \sum_{s=1}^{S} \log\int p(X_{s}|E)p(E)dE
   \label{eqt8}
   \end{equation}
   Continuing the theory of variational Bayes, we regard $E$ as the latent 
   variable in variational autoencoder (VAE)~\cite{kingma2022autoencoding}
    and use variational 
   posterior $q(E|X)$ to approximate the intractable posterior 
   $p(E|X)$, thus the evidence lower bound (ELBO) reads: 
   \begin{equation}
    \mathcal{L}^{s}_{ELBO}=-KL(q(E|X_{s})||p(E))+E_{q(E|X_{s})}[\log p(X_{s}|E)]
    \label{eqt9}
    \end{equation}
  From the VAE view, for each sample $X_{s}$, the inference model encodes it 
  to output latent variable $E$ with $q(E|X_s)$, and then the decoder 
  reconstructs $X_{s}$ from $E$ via $p(X_{s}|E)$. From the causal graph 
  view, defining that $\mathcal{X}$ represents the domain of samples: 
  $\{X_s\}^{S}_{s=1} \in \mathcal{X}$, $\mathcal{E}$ represents the 
  domain of exogenous variable $E$, $\widehat{\mathcal{X}}$ represents 
  the domain of causal representation. Therefore, a variational model 
  consists of two functions: an exogenous variable inference: 
  $\mathcal{X} \rightarrow \mathcal{E}$, and observation relation 
  reconstruction: $\mathcal{E} \rightarrow \widehat{\mathcal{X}}$.
\end{itemize}
The shortcoming of this framework is that when we extend it to 
indefinite data with $M$ different skeletons 
$\mathcal{G}_{m}=\{\mathcal{E}_{m}, \mathcal{V}_{m}\}^{M}_{m=1}$, 
the variational posterior $q(E|X_{m,s})$ must be optimized separately 
for each of them. Meanwhile, 
for the $E_{m} \in \mathbb{R}^{N \times D} $ in each skeleton, it is 
hard to define the prior $p(E_{m})$ with the brief distribution. DAG-GNN
~\cite{yu2019dag} modeled it as the standard matrix normal 
$p(E_{m})=\mathcal{M}\mathcal{N}_{N \times D}(0,I,I)$ to apply to 
variables of multi-independent values, but there is always 
a correlation between any two values in indefinite data 
(and it also faced building a range of models).  
Here we ascribe the failure on multi-value data to the inability to 
make each value independent. As a result, 
variational models for indefinite data from this framework cannot 
1) utilize enough samples to discover one skeleton, 
2) use one model to finish all samples, 
3) find an adequate distribution assumption for prior. 

\section{Dynamic Variational Inference Model}
\subsection{Causal Strength as Latent Variables}
Variational models employed for causal discovery 
are based on the structural causal models (SCMs) 
(The definition of SCMs is provided in Appendix A.3). 
Conventionally, noise terms are treated as latent variables 
within these variational formulations, 
enabling the assignment of an i.i.d. prior distribution. 
We offer a detailed explanation of this proposed framework 
in Section~ref{AppC}, and analyse the challenges when 
deployed on multi-skeleton and multi-value data. 

Therefore, we designed a probabilistic framework 
utilizing causal strength as latent variables to 
achieve these challenges. Due to the trait of multi skeletons, 
the variables (nodes) are different and incapable of satisfying the 
same distribution when samples belong to different skeletons. 
However, a set of causal graphs satisfy a particular distribution 
in which one skeleton's causal strength distributionally resembles those of 
other skeletons, even though they differ. We can 
take advantage of such variance induced by many skeletons to learn 
causal graphs without refitting a new model. Meanwhile, the causal strength 
is single-value, so we do not concern about how to assume the distribution  
of multi values.

Given an indefinite data $\{X_s\}^{S}_{s=1}$, our framework obeys the 
CAM from Equation 3 in Appendix. A.3, and we can build the same 
SCM to describe the causal discovery of indefinite data: 

\begin{equation} 
    E=f_{2}((I-A)f_{1}(X)) 
    \label{eqt10}
\end{equation}
\begin{equation} 
    X=f_{4}((I-A)^{-1}f_{3}(E)) 
    \label{eqt11}
\end{equation}
where $X$ and $E$ are both multi-value data, but $A$ is still a 
single-value matrix representing the stable statistic of samples 
from one skeleton, 
$A\in \mathbb{R}^{N \times N}, X \in \mathbb{R}^{N \times D}, E \in \mathbb{R}^{N \times D}$. 
In traditional causal discovery, acyclicity constraints 
are often introduced to ensure that the adjacency matrix is causal. 
However, given Hypothesis 1 in Appendix A.2, 
we merely need to apply a strict triangular mask to the adjacency matrix.

From the overall view, sampling from a set 
of DAGs $\mathcal{G}_{m}=\{\mathcal{E}_{m}, \mathcal{V}_{m}\}^{M}_{m=1}$ 
is equal to generating a set of causal strengths which reads: 
\begin{equation}
  p(A)=\{p(A_m)\}^{M}_{m=1}
\label{eqt12}
\end{equation}
From the generation model view, we can briefly assume that 
$I-A = f_{5} (X)$ to represent the encoder that takes $X$ as input 
and return $I-A$ and thus we can obtain: 
\begin{align}
  E=f_{2}((I-A)f_{1}(f^{-1}_{5}(I-A)))=f_{2\circ 1\circ 5}(I-A)
\label{eqt13}
\end{align}
Equation~\ref{eqt13} justifies that there exists an equal generation 
function replacing $p(E)$ with $p(I-A)$. Figure~\ref{figtwomodel} 
differs our framework from the existing framework in architecture. 
In the causal view, our framework consists of two functions: a causal 
strength encoder: $\mathcal{X} \rightarrow \mathcal{G}$ and a 
relation reconstruction decoder: 
$\mathcal{G}\times \mathcal{E} \rightarrow \widehat{\mathcal{X}}$.
Consequently, we deem $W=I-A$ as latent variables and 
use variational posterior $q(W|X)$ instead of $q(E|X)$ to 
describe the new log-evidence: 
\begin{equation}
\begin{split}
  &\frac{1}{M}\frac{1}{S} \sum_{m=1}^{M}\sum_{s=1}^{S} \log p(X_{s,m})=\frac{1}{M}\frac{1}{S} \sum_{m=1}^{M}\sum_{s=1}^{S} \log\int p(X_{s,m}|W)p(W)dW
  \label{eqt15}
\end{split}
\end{equation}
When $M=1$, Equation~\ref{eqt15} resembles Equation 9 in Appendix B, indicating 
that multi-skeleton maximization of log-evidence is the sum of 
maximization of log-evidence from each skeleton, which corroborates the 
superiority of our framework being capable of utilizing one model to 
learn all causal graphs from indefinite data. 

\begin{figure*}
    \centering
    \subfigure[DAG $\mathcal{G}$]{
      \includegraphics[width=0.2\textwidth]{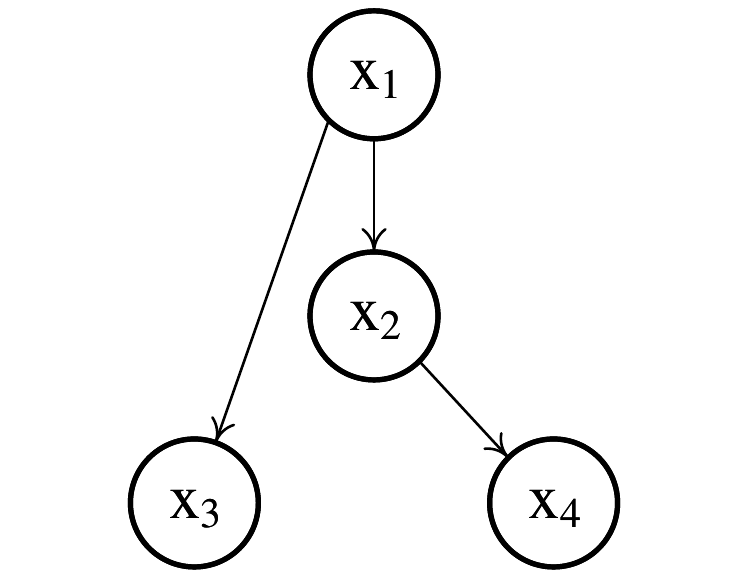}}
    \subfigure[DAG $\mathcal{H}$]{
      \includegraphics[width=0.2\textwidth]{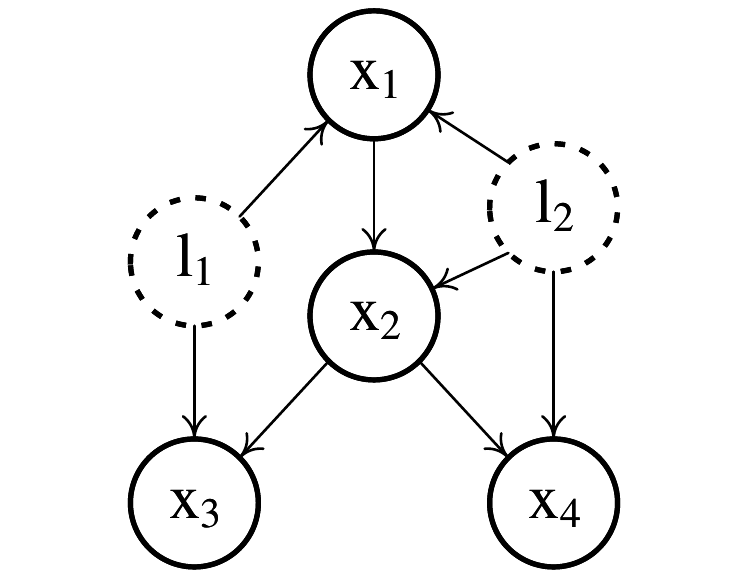}}
    \subfigure[DAG $\mathcal{O}$]{
      \includegraphics[width=0.2\textwidth]{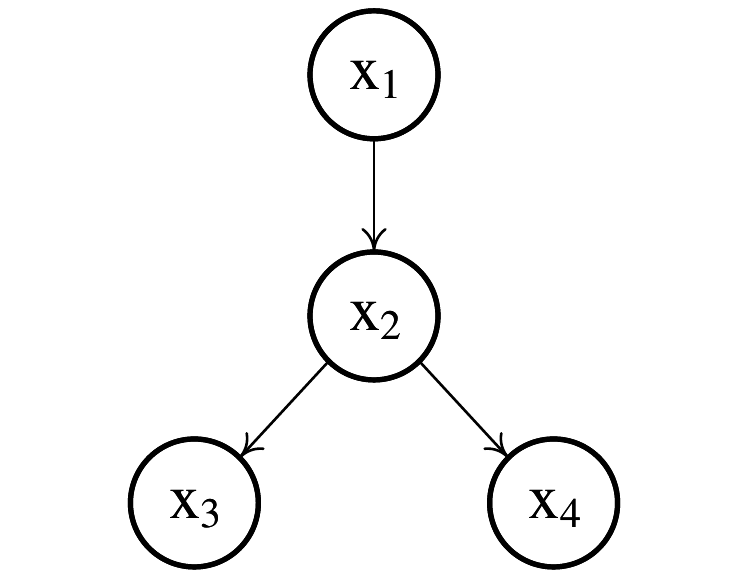}}
    \subfigure[DAG $\mathcal{C}$]{
      \includegraphics[width=0.2\textwidth]{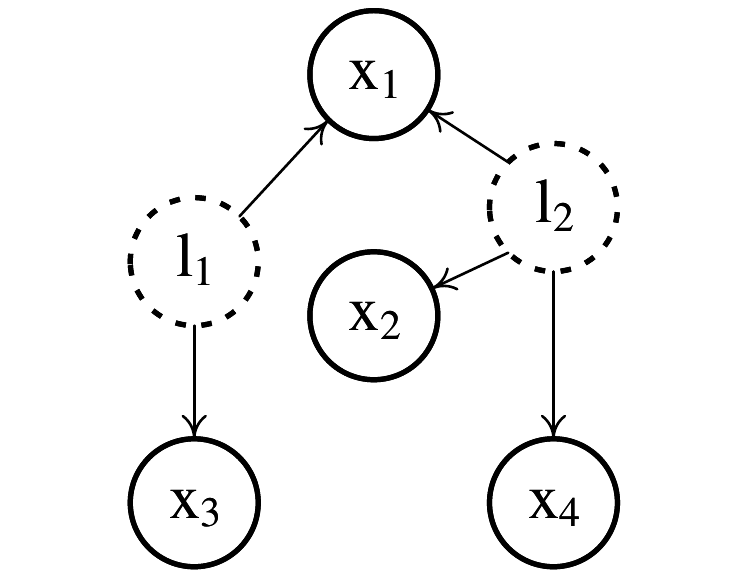}}
    \caption{Four causal DAGs discussed in this paper. $\mathcal{G}$ 
    represents the wrong causal structure lacking the consideration of 
    latent confounders, $\mathcal{H}$ represents the true causal 
    structure given the whole observed variables and latent confounders, 
    $\mathcal{O}$ and $\mathcal{C}$ are two subgraphs disentangled from 
    $\mathcal{H}$. $\mathcal{O}$ has and only has relations between observed 
    variables (i.e., $x_{1} \to x_{2}$), and $\mathcal{C}$ has and only 
    has relations from latent confounders to observed variables (i.e., 
    $l_{1} \to x_{1}$). }
    \label{figdisentanglement}
  \end{figure*}

Corresponding for the ELBO, the KL term and reconstruction term need 
to be modified to : 
\begin{equation}
\begin{split}
  \mathcal{L}^{s,m}_{ELBO}=&-KL(q(W|X_{s,m})||p(W))+E_{q(W|X_{s,m})}[\log p(X_{s,m}|W)] 
  \label{eqt16}
\end{split}
\end{equation}
For simplicity, we follow the typical setting in definite data and 
model the prior as the standard normal 
$p(I-W)=\mathcal{M}\mathcal{N}_{N \times N}(0,I,I)$, which indicates 
that each causal strength $p(f_{ij})=\mathcal{N}(0,1)$. Note that 
even though the nodes are probably connected in a true graph, however, 
they are independent in prior.

By elucidating the causal strengths from different skeletons 
within a particular distribution, we can utilize variational inference 
to separate the causal relation prediction from the learning of 
observed representation. This setup allows us to ensemble all samples 
and learn the invariance of diverse skeletons via a single model, 
significantly improving sample utilization. 

Besides, we resort to a single-value matrix format to represent the causal 
strengths between all nodes, which makes it amenable to the assumption of a prior 
distribution. Consequently, we replace the 
estimation of independent noise with the estimation of causal strength 
to achieve a deductive lower bound shift, especially in the case of 
multi-valued data, where independent noise distributions are not 
easily assumed.

\subsection{Confounding Disentanglement}~\label{cd}
Considering the confounders is another breakthrough of our method, 
especially for indefinite data. From the probabilistic framework, 
a causal structure can be modeled by a causal DAG, 
where the nodes are observed variables, 
and the edges represent the direct causal strength these 
variables have on one another. However, in many realistic circumstances, 
only a part of the variables are observed, i.e., 
the latent confounders on 
observed variables is induced. 
Therefore, in the presence of latent confounders, the SCM needs to be 
added a confounding variable. 
\begin{definition}[SCM including confounders]
  An SCM is a 3-tuple $\langle V_{m},\mathcal{F}_{m},\mathbb{P}_{m}\rangle $, where 
  $V_{m}=X_{m}\cup L_{m}$ is the total set of observed variables 
  $X_{m}=\{x_{m,i}\}^{S}_{i=1}$ and latent confounding variables 
  $L_{m}=\{l_{m,i}\}^{S}_{i=1}$. Structural equations 
  $\mathcal{F}_{m}=\{f_{m,i}\}^{S}_{i=1}$ are functions that determine 
  $X$ with $x_{m,i}=f_{m,i}(Pa(x_{{m,i}}),l_{m,i})$, 
  where $Pa(x_{{m,i}})\subseteq X_{m}$. 
  $\mathbb{P}_{m}(X)$ is a distribution over $X_{m}$. 
  \label{def4}
\end{definition}
Definition~\ref{def4} describes two relations in a causal DAG under 
confounding: the relation between observed variables 
and the relation from confounders to observed variables. 
Corresponding to the CAM assumption, we add a confounding 
term to indicate this effect: 
\begin{equation} \
\begin{split}
  x_{m,j}=&\sum_{x_{m,i}\in Pa(x_{m,j})} f_{m,ij} x_{m,i}+\sum_{l_{m,k}\in Ec(x_{m,j})} g_{m,kj} l_{m,k} +\epsilon_{x_{m,j}}
  \label{eqt17}
\end{split}
\end{equation}
where $Ec(x_{m,j})$ is the confounder set having effects on $x_{m,j}$. 
$K$ is the number of latent confounders. 
$x \in \mathbb{R}^{N \times D},l \in \mathbb{R}^{K \times D},\epsilon_{x_{s,j}} \in \mathbb{R}^{N \times D}, 0\leq i \neq j,k < N$. 
Besides, the linear SEM model also modifies: 
\begin{equation}
 X=AX+BL+E
 \label{eqt19}
\end{equation}
Equations~\ref{eqt17} and~\ref{eqt19} indicate that latent confounders 
are a critical problem in our research: when they exist, 
non-autoregression SEM invalidates the VAE. 
Hence, to eliminate the 
effect of confounders and reconstruct the true causal relations, 
we consider the following disentanglement model in this paper: 
\begin{align}
  \mathcal{H}&=\mathcal{O} \cup \mathcal{C}
    \label{eqt20}
\end{align}
where $\mathcal{H}=\{(I-A)^{-1}(BL+E),\mathcal{E}_{\mathcal{H}}\}$,  
         $\mathcal{O}=\{(I-A)^{-1}E,\mathcal{E}_{\mathcal{O}}\}$, 
         $\mathcal{C}=\{(I-A)^{-1}BL,\mathcal{E}_{\mathcal{C}}\}$.
From a causal graph view, 
graph $\mathcal{O}=\{X, \mathcal{E}_{\mathcal{O}}\}$, 
$\mathcal{E}_{\mathcal{O}}$ represents the edge 
`$x_{i} \to x_{j}$', and graph 
$\mathcal{C}=\{X\cup L, \mathcal{E}_{\mathcal{C}}\}$. 
$\mathcal{E}_{\mathcal{C}}$ represents the edge 
`$l_{k} \to x_{j}$'. Graph $\mathcal{H}$ is the full causal 
graph with all observed and latent relations. Note that 
$\mathcal{H}$ is only in theory because we can not obtain 
confounders. $\mathcal{H}=\mathcal{G}$ 
if we omit the latent confounders, which embraces the 
traditional causal discovery aspect (Equation 4 in Appendix B).
$\mathcal{H}=\mathcal{O}$ 
if there are no confounders in this causal skeleton.
See also Figure~\ref{figdisentanglement} for an illustration. 

In general, the ultimate goal of causal discovery under confounding 
is to correct $\mathcal{G}$ and recover $\mathcal{H}$ from observed 
variables set $X$. However,  indefinite data, such a complicated data 
paradigm, makes it stubborn to achieve $\mathcal{H}$ because some 
prevalent assumptions about solving confounding can not hold here 
(e.g., there is no relation between any two observed variables
~\cite{squires2022causal}, or all observed variables are affected by 
one confounder~\cite{agrawal2021decamfounder}), which results in the 
problems that we can not know and assume the locations, numbers, and 
effects of confounders. 

\begin{figure*}
    \includegraphics[width=1\linewidth]{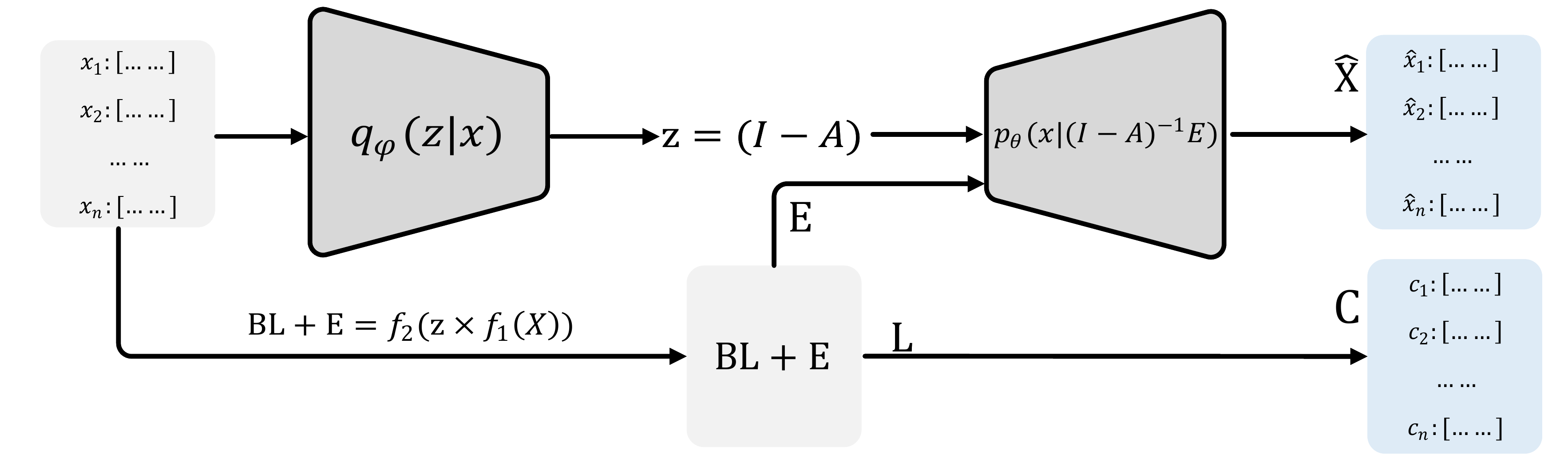}
    \caption{A probabilistic implementation of biCD. 
    $q_{\varphi}(z|\mathcal{X})$ predicts the causal strength weight from 
    the input $x$. The predicted latent variable $z=(I-A)$ in the 
    SEM matrix, and then an observed variable decoder 
    $p_{\theta}((x|(I-A)^{-1}E))$ learns to predict 
    $\widehat{X}$ given the disentangled $E$ 
    and inverse of predicted $z$.}
    \label{figbiCD}
  \end{figure*}

To this end, we design this causal disentanglement, which makes 
relations of observed variables amenable without assuming  
confounders. For $\mathcal{O}$, we follow the variational interference 
in Equation~\ref{eqt11}, 
which can reconstruct the deconfounding variables $\widehat{X}$. 
For $\mathcal{C}$, we would like to compute the confounding effects $C_{i}$
on each observed variables $X_{i}$ instead of the value of confounders. 
The confounding effects $C \in \mathbb{R}^{N \times D}$ structurally 
resemble reconstruction variables $\widehat{X} \in \mathbb{R}^{N \times D}$. 
From the CAM view, it reads: 
\begin{equation} 
\begin{split}
  &c_{m,j}=\sum_{l_{m,k}\in Ec(x_{m,j})} g_{m,kj} l_{m,k} \\
  &(c \in \mathbb{R}^{N \times D}, l\in \mathbb{R}^{K \times D}, 0\leq i \neq j,k < N)
  \label{eqt24}
\end{split}
 \end{equation}
Due to the assumption of causal strength is irrelevant to 
`confounding strength' $g$, equation~\ref{eqt24} is 
intractable without approximate statistics as we shown in following, 

\begin{definition}
  there exist an approximate estimation 
  \begin{equation}
    C_{j}=\frac{\mathbb{P}(x_{j}) \mathbb{P}(L|x_{j})}{\sum_{i}^{N} \mathbb{P}(x_{i}) \mathbb{P}(L|x_{i}) } x_{j} 
    \label{eqt26}
\end{equation}
  when the latent variables $l$ vastly outweigh independent noise $\epsilon_{x}$. 
  \label{def5}
\end{definition}

Definition~\ref{def5} (Proofs are shown in Appendix C) describes an expectation statistic irrelevant to 
$B$ while involved with $E$, which collaborates the inductive bias that 
when confounding effects drastically exceed independent noise, $X$ is 
approximately contributed by $L$ rather than $E$. We thus design  
a dynamic reconstruction loss $l_{r}$: when $l\gg \epsilon_{x}$ (i.e., the 
confounding effects are significant), $l_{r}$ measures the distance 
between $X$ and $\widehat{X}+C$; on the contrary, in the case 
that confounding effects are negligible, $l_{r}$ measures the 
distance between $X$ and $\widehat{X}$ as well as $C$ is hard to 
estimate.  

\subsection{A Probabilistic Implementation}\label{api}

The causal strength model contributes an available variation posterior 
$q((I-A)|X)$, and confounding disentanglement contributes a dynamic 
reconstruction loss. In this section, 
we will introduce the probabilistic implementation. 

We formalized dynamic variational inference model as follows. 
$\mathcal{X}$ represents the domain of 
data samples: $\{X_{s}\}^{S}_{s=1}\in \mathcal{X}$. $\mathcal{O}$ 
represents the domain of all possible causal strengths on 
$\mathcal{X}$. Note that the effect of latent confounders (i.e., 
$l_{k} \to x_{i}$) is disentangled in $\mathcal{O}$. 
Therefore, our method consists of three functions: a causal graph encoder 
$f_{\varphi}: \mathcal{X} \to \mathcal{O}$, an observation relation 
decoder $f_{\theta}:\mathcal{O}\times \mathcal{E} \to \widehat{\mathcal{X}}$, 
and an estimation function 
$f_{\delta}: C=f_{\delta}(\mathbb{P}(x), \mathbb{P}(L|x))$. 

In Figure~\ref{figbiCD}, we resort to VAE to design the functions 
$f_{\varphi}$ and $f_{\theta}$. Specifically, we designed a gnn-based 
encoder $q_{\varphi}(z|X)$ to output a distribution over $z$. 
Intrinsically, $z$ represents the causal strength of observed 
variables, also called the edge in causal subgraph 
$\mathcal{O}$. Hence, the decoder $p_{\theta}((x|(I-A)^{-1}E))$ 
probabilistically learns to model the invariance of the various causal 
graph, outputting the causal representation 
$\widehat{X}$ without confounding. 

\subsubsection{Encoder}
The encoder $q_{\varphi}(z|\mathcal{X})$ applies a graph attention 
module $f_{att,\varphi}$~\cite{velivckovic2017graph} to the input. 
It produces an adjacent matrix across a causal skeleton mask. (We 
regulate the mask as a DAG satisfying the requirement of $\mathcal{O}$)
\begin{equation}
q_{\varphi}(z|\mathcal{X})=softmax(f_{att,\varphi}(X))
\label{eqt27}
\end{equation}
We introduced a Gumbel distributed  noise $e$
~\cite{maddison2016concrete} to be capable of 
backpropagating via the discrete distribution samples. 
\begin{equation}
  z \sim softmax((I-A)+e)
  \label{eqt28}
  \end{equation}
The output $z$ implies the possible distribution of causal strength 
over $\mathcal{X}$. Specifically, $z_{i,j}=1$ indicates a 
high probability relation $x_{j} \to x_{i}$. 
Note that we set $A$ a strict lower triangle matrix to hold the acyclic 
graph. 
In this cohesive way, our 
method addresses multi-value and multi-skeleton challenges.

\subsubsection{Decoder}
First, we extract $E$ by utilizing a multi-layer 
perceptron (MLP) upon graph $\mathcal{O}$:
\begin{align}
  BL+E=&GNN_{enc}(f_{att,\varphi}(X), X)\\
  E = &MLP_{E}(BL+E)
  \label{eqt29}
  \end{align}
where $GNN_{enc}$ is instantiated by graph neural network: 
$GNN(\mathsf{A}  ,\mathsf{X} )=eLU(\mathsf{A} \times (\mathsf{X} \times \mathsf{W}) )$, 
which yields a nonlinear multiple of adjacent matrix $\mathsf{A} \in \mathbb{R}^{N \times N}$, 
feature matrix $\mathsf{X}  \in \mathbb{R}^{N \times D}$ 
and weight matrix $\mathsf{W}_{enc}  \in \mathbb{R}^{D \times H}, 
\mathsf{W}_{dec}  \in \mathbb{R}^{H \times D}$. 
Corresponding to Equation~\ref{eqt6}, the SEM encoder should be: 
$BL+E=f_{2}(z\times f_{1}(X))$ in which a nonlinear function 
$f_{1}$ is set for $X$. We do adopt a dropout layer to initialize it: 
$X=dropout(X)$, while we omit it and replace it by `$X$' for simplicity.  
Then, the decoder accumulated the incoming messages to each node via 
causal strength $z$ and employed a new graph neural network $GNN_{dec}$: 
  \begin{align}
    p_{\theta}((\hat{x}|z^{-1}E))=GNN_{dec}(z^{-1},E)
    \label{eqt30}
    \end{align}
The output of the decoder $\widehat{X} \in \mathbb{R}^{N \times D}$ equals 
the dimension of $\mathcal{X}$ and  it is the pure representation 
of $X$ without confounding.

\subsubsection{Confounding Effect Estimation}
We used the same MLP module to extract $L$ and two sigmoid functions: 
$\sigma_{\mathbb{P}(x_{j})}(\cdot )$ and $\sigma_{\mathbb{P}(L|x_{j})}(\cdot )$, 
to project $\mathbb{P}(x_{j})$ and $\mathbb{P}(L|x_{j})$ into the range 
of $(0,1)$, which expresses the probability estimating $C_{j}$. 
\begin{align}
  &L = MLP_{L}(BL+E)\\
  &C_{j}=\frac{\sigma_{\mathbb{P}(x_{j})}(x_{j} ) \sigma_{\mathbb{P}(L|x_{j})}(L|x_{j})}{\sum_{i}^{N} \sigma_{\mathbb{P}(x_{i})}(x_{i} ) \sigma_{\mathbb{P}(L|x_{i})}(L|x_{i}) } x_{j} 
  \label{eqt32}
  \end{align}
The output $C \in \mathbb{R}^{N \times D}$ of Estimation module equals 
the individual-specific effects of confounding on each $x_{j}$ if there 
exactly exists strong confounding, we will detail this problem in 
following. 

\begin{table*}[htbp]
  \footnotesize
  \centering
  \resizebox{\textwidth}{!}{
    \resizebox{0.95\linewidth}{!}{
    \subtable[Results in Sample=$\{5, 10, 50\}$]{
    \begin{tabular}{|c|c|c|c|}
      \hline
    \multirow{2}{*}{Method}&\multicolumn{3}{c}{\textbf{Sample}}\\
    \cline{2-4}
    &Syn1&Syn2&Syn3\\
    \hline
    GIN&0.504$_{\pm 0.005}$&0.512$_{\pm 0.006}$&0.541$_{\pm 0.004}$\\
    LFCM&0.512$_{\pm 0.009}$&0.518$_{\pm 0.006}$&0.545$_{\pm 0.007}$\\
    pcss&0.537$_{\pm 0.004}$&0.546$_{\pm 0.003}$&0.582$_{\pm 0.004}$\\
    DAG-GNN&0.551$_{\pm 0.009}$&0.567$_{\pm 0.010}$&0.585$_{\pm 0.009}$\\
    \hline
    biCD&\textbf{0.581}$_{\pm 0.003}$&\textbf{0.595}$_{\pm 0.002}$&\textbf{0.623}$_{\pm 0.003}$\\
    \hline
  \end{tabular}}}

  \resizebox{0.95\linewidth}{!}{
  \subtable[Results in Confounders=$\{1, 5, 10\}$]{
    \begin{tabular}{|c|c|c|c|}
      \hline
    \multirow{2}{*}{Method}&\multicolumn{3}{c}{\textbf{Confounder}}\\
    \cline{2-4}
    &Syn3&Syn4&Syn5\\
    \hline
    GIN&0.541$_{\pm 0.004}$ &0.543$_{\pm 0.006}$&0.558$_{\pm 0.003}$\\
    LFCM&0.545$_{\pm 0.007}$&0.540$_{\pm 0.008}$&0.555$_{\pm 0.003}$\\
    pcss&0.582$_{\pm 0.004}$&0.553$_{\pm 0.003}$&0.541$_{\pm 0.004}$\\
    DAG-GNN&0.585$_{\pm 0.009}$&0.588$_{\pm 0.012}$&0.571$_{\pm 0.021}$\\
    \hline
    biCD&\textbf{0.623}$_{\pm 0.003}$&\textbf{0.619}$_{\pm 0.005}$&\textbf{0.611}$_{\pm 0.003}$\\
    \hline
  \end{tabular}}}}

  \resizebox{\textwidth}{!}{
    \resizebox{0.95\linewidth}{!}{
  \subtable[Results in Observed nodes=$\{20, 50, 100\}$]{
      \begin{tabular}{|c|c|c|c|}
        \hline
    \multirow{2}{*}{Method}&\multicolumn{3}{c}{\textbf{Observed node}}\\
    \cline{2-4}
    &Syn6&Syn3&Syn7\\
    \hline
    GIN&0.578$_{\pm 0.004}$&0.541$_{\pm 0.004}$&0.516$_{\pm 0.007}$\\
    LFCM&0.588$_{\pm 0.007}$&0.545$_{\pm 0.007}$&0.536$_{\pm 0.001}$\\
    pcss&0.611$_{\pm 0.005}$&0.582$_{\pm 0.004}$&0.552$_{\pm 0.003}$\\
    DAG-GNN&0.621$_{\pm 0.018}$&0.585$_{\pm 0.009}$&0.588$_{\pm 0.017}$\\
    \hline
    biCD&\textbf{0.647}$_{\pm 0.004}$&\textbf{0.623}$_{\pm 0.003}$&\textbf{0.626}$_{\pm 0.002}$\\
    \hline
  \end{tabular}}}

  \resizebox{0.95\linewidth}{!}{
  \subtable[Results in Pervasiveness=$\{0.1, 0.4, 0.7\}$]{
      \begin{tabular}{|c|c|c|c|}
        \hline
    \multirow{2}{*}{Method}&\multicolumn{3}{c}{\textbf{Pervasiveness}}\\
    \cline{2-4}
    &Syn8&Syn9&Syn3\\
    \hline
    GIN&0.572$_{\pm 0.005}$&0.559$_{\pm 0.002}$&0.541$_{\pm 0.004}$\\
    LFCM&0.581$_{\pm 0.007}$&0.569$_{\pm 0.002}$&0.545$_{\pm 0.007}$\\
    pcss&0.545$_{\pm 0.002}$&0.559$_{\pm 0.003}$&0.582$_{\pm 0.004}$\\
    DAG-GNN&0.577$_{\pm 0.012}$&0.577$_{\pm 0.013}$&0.585$_{\pm 0.009}$\\
    \hline
    biCD&\textbf{0.633}$_{\pm 0.007}$&\textbf{0.635}$_{\pm 0.005}$&\textbf{0.623}$_{\pm 0.003}$\\
    \hline

  \end{tabular}}}}
  \caption{AUROC for causal graph on synthetic dataset, 95$\%$ 
  confidence interval shown. }
  \label{tabauroc}
\end{table*}

\subsubsection{Dynamic Reconstruction Error}
Unlike definite data, the indefinite data w.r.t. different 
causal skeletons are unlikely to have sufficient confounding 
effects in all causal samples. For those skeletons without confounding or with 
negligible confounding, graph $\mathcal{C}$ is meaningless. 
The reconstruction error $l_{rc}$ (We adopt mean squared error (MSE) in implementation)
only involved $X$ and $\widehat{X}$, so our reconstruction error is : 
\begin{align}
l_{rc}(X,\widehat{X})=E_{q_{\varphi}(z|\mathcal{X})}[MSE(x,f_{\theta}(z^{-1}E)]
\label{eqt34}
\end{align}  
However, for those skeletons with obvious confounding, the reconstruction error 
must consider the confounding effects $C$. The reconstruction error 
$l_{rc}$ involved $X$, $\widehat{X}$, and $C$. 
\begin{align}
  l_{rc}(X,\widehat{X}+C)=E_{q_{\varphi}(z|\mathcal{X})}[MSE(x,(f_{\theta}(z^{-1}E)+C)]
  \label{eqt37}
  \end{align} 
Therefore, we naturally design a confounding score for each graph as 
$\omega(L)=rank(L)/N$ ($L$ is computed by equation~\ref{eqt32}. 
The graphs with high $\omega(L)$ can be regarded as confounding 
samples because the high rank of the matrix $L\in \mathbb{R}^{N \times D}$ 
stands for the extensive independent noises in $L$, which indicates 
that sufficient exogenous confounding variables point to the $X$, and 
vice versa. Finally, the reconstruction error and ELBO can be encapsulated by: 
\begin{align}
  &l_{rc}=\omega(L) l_{rc}(X,\widehat{X}+C)+ (1-\omega(L))l_{rc}(X,\widehat{X})\\ 
  &\mathcal{L}=l_{rc}-KL[q_{\varphi}(z|\mathcal{X})||\mathbb{P}(z)]
  \label{eqt40}
  \end{align}

\section{Experiments} \label{exp}

In this section, we conduct extensive experiments to answer two 
research questions:
\textbf{RQ1}:How effective is our method in causal discovering from 
multi-skeleton and multi-value data?
\textbf{RQ2}:What are the insights of $\mathcal{O}$ and $\mathcal{C}$? 
Especially, how to justify the confounding disentanglement? 

\subsection{Datasets and Experiment Setup} 
To show the efficacy of our approach, we applied it to both synthetic and 
real-world data. In the synthetic data, we construct four sets of datasets 
to investigate four key ingredients respectively: the number of 
\textbf{samples} (5, 10, 50), the number of \textbf{confounders} (1, 5, 10), 
the number of \textbf{observed variables} (20, 50, 100), and the 
\textbf{pervasiveness} (0.1, 0.4, 0.7) of confounding. 
Besides, for real-world data, we conducted 
experiments on affective reasoning datasets, including 
emotion-cause conversation data. We introduced these datasets, 
baselines adopted, and experiment setup in Appendix E and F.

The synthetic datasets function as semi-definite data 
which seeks to evaluate the capabilities under multi skeletons: 
we measure predicted causal graph performance by 
the area under the receiver operator curve (AUROC) of predicted causal 
strength probabilities over test samples. We compare the MSE loss 
of $C$ between the estimation of the above deconfounding approaches 
and the ground truth value. 
The real-world datasets, composed of indefinite data, aim to measure 
the actual enhancement provided by our 
causal representations for high-level tasks via F1 scores.

\subsection{Synthetic Data}
\subsubsection{Main Results (RQ1)}

\begin{figure*}
  \centering
  \subfigure[$\#$Sample]{
    \includegraphics[width=0.23\linewidth]{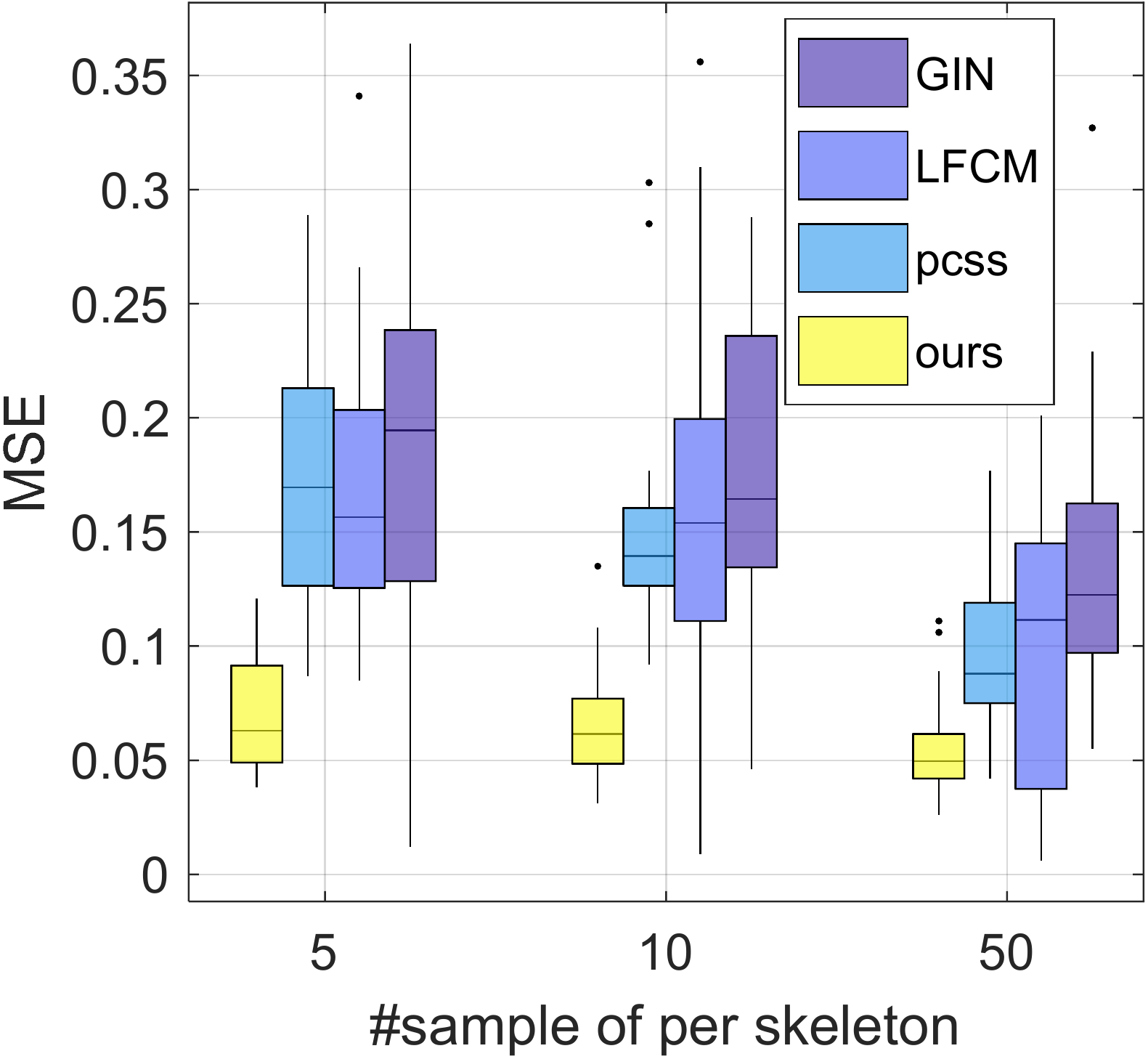}}
  \subfigure[$\#$Confounder]{
    \includegraphics[width=0.23\linewidth]{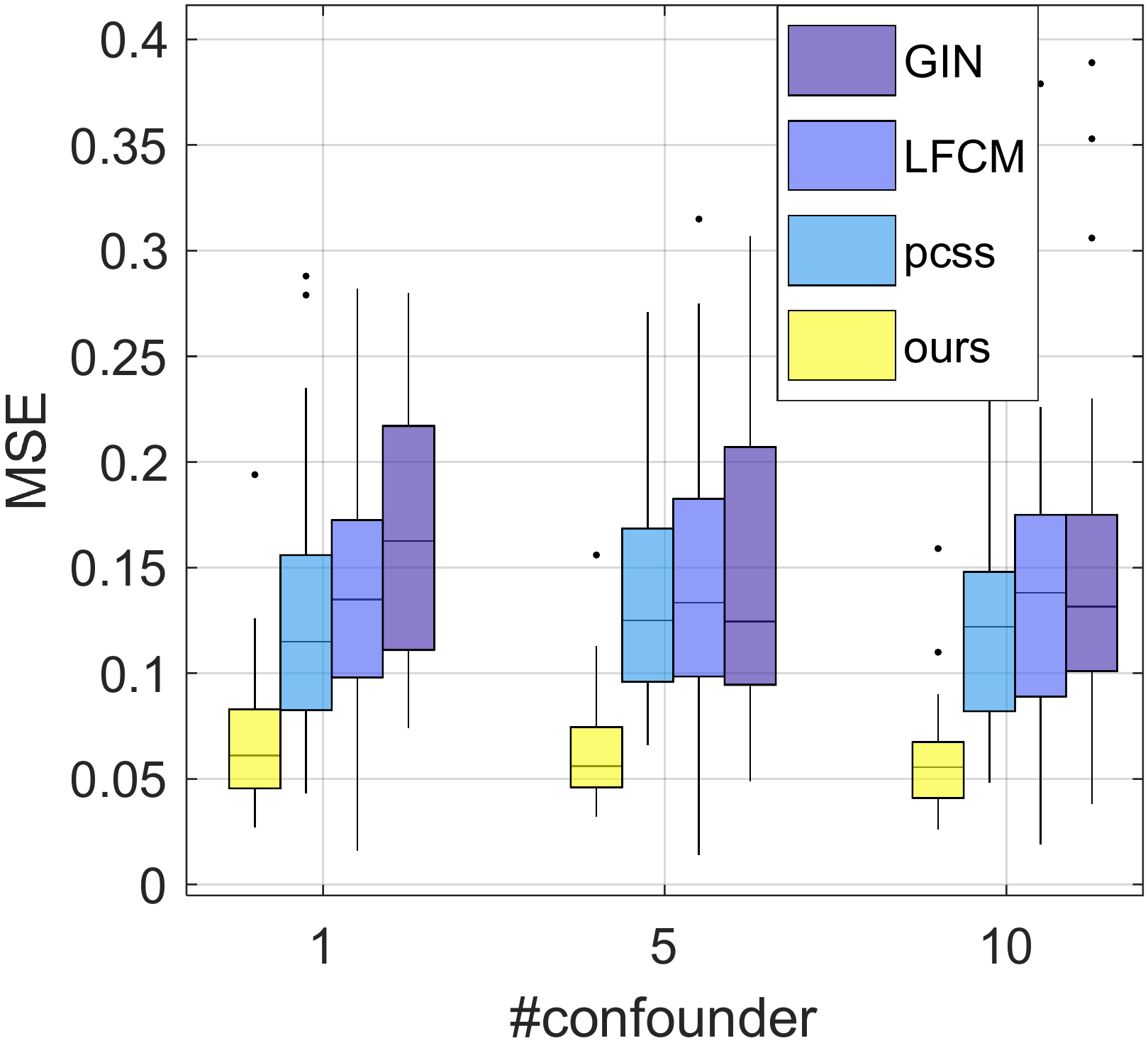}}
  \subfigure[$\#$Observed nodes]{
    \includegraphics[width=0.23\linewidth]{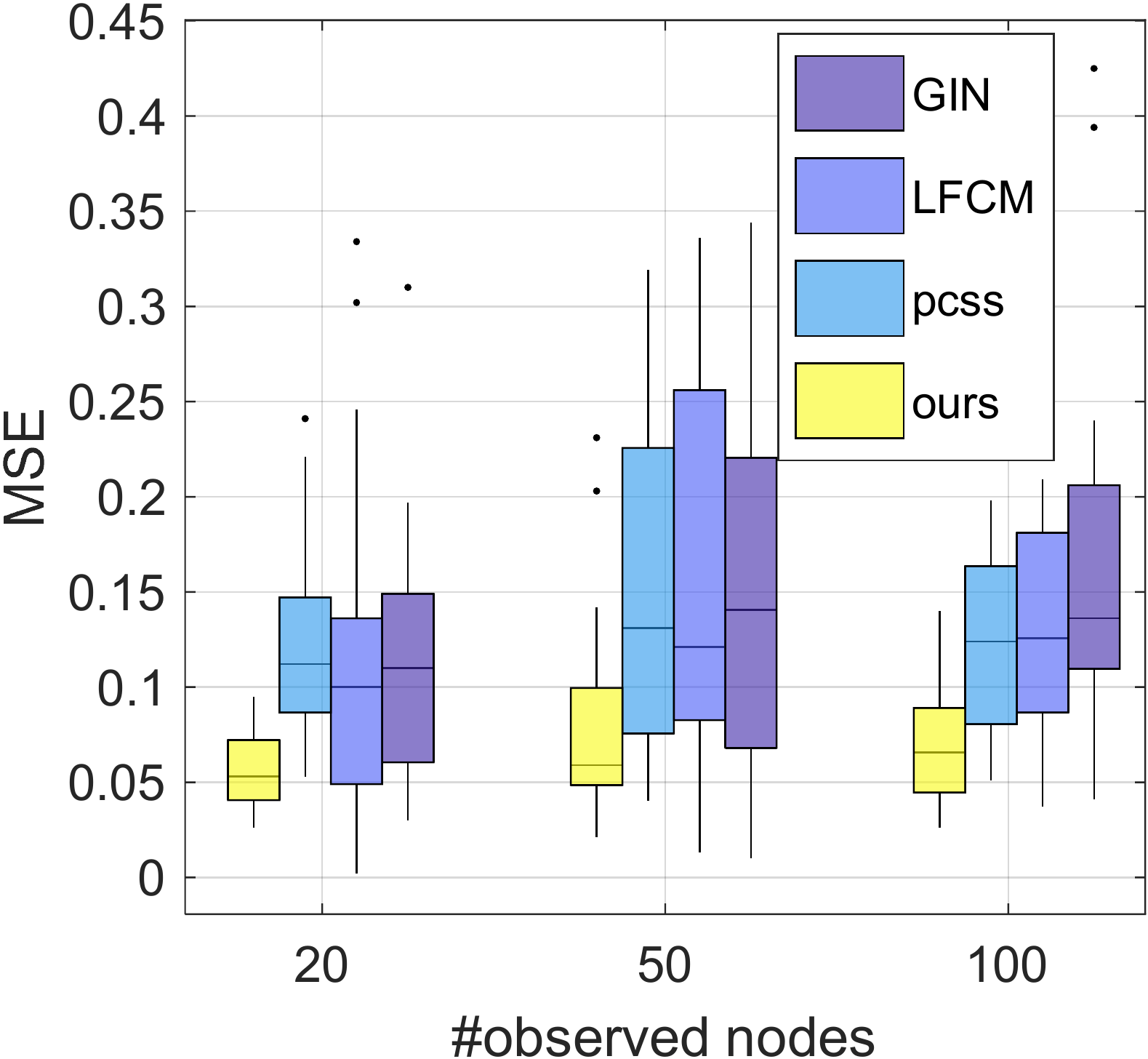}}
  \subfigure[Pervasiveness]{
    \includegraphics[width=0.23\linewidth]{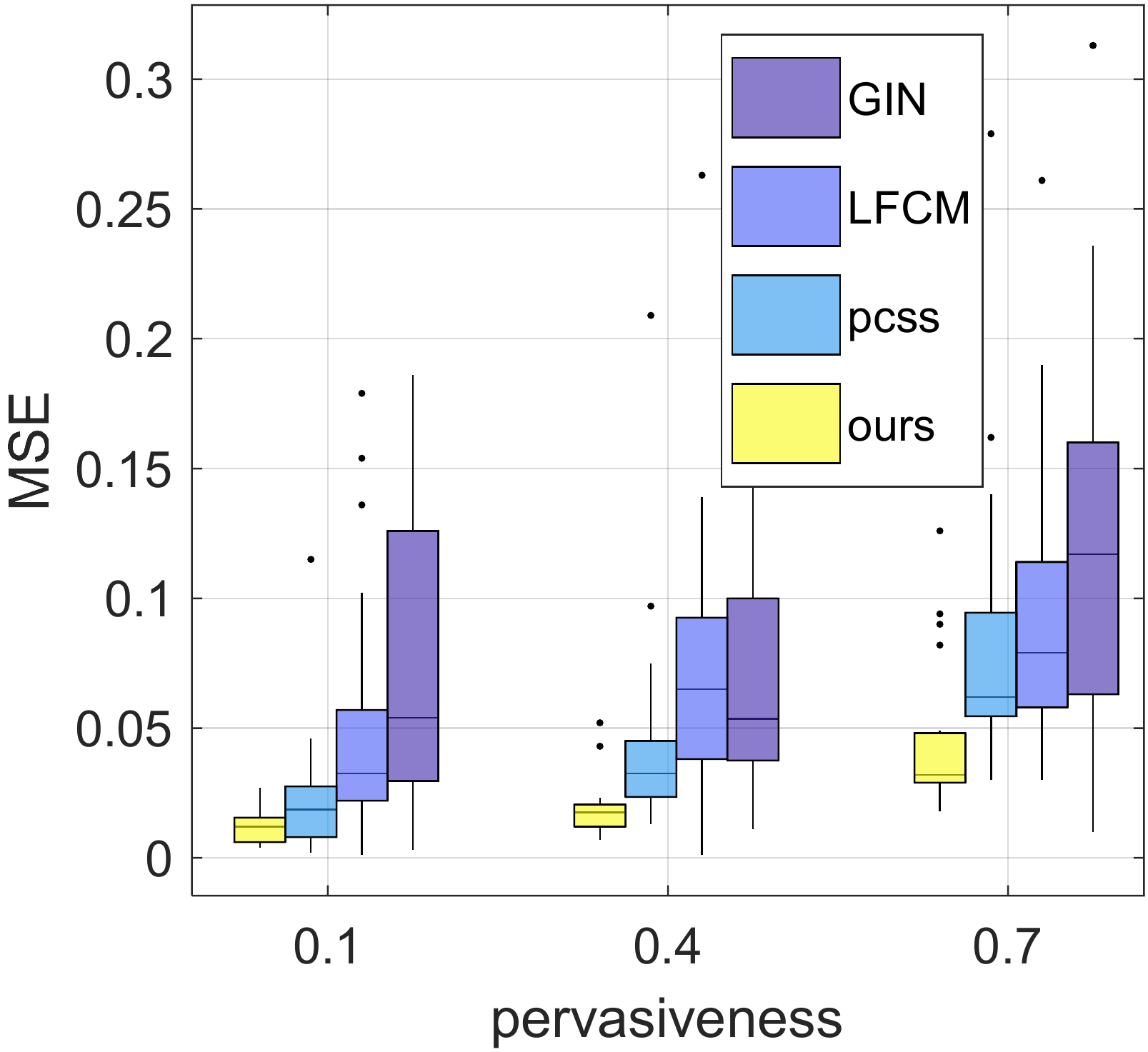}}
  \caption{MSE error across all ingredients setting for estimating 
  $C$ via GIN, LFCM, pcss, and ours.}
  \label{figmse}
\end{figure*}
First, we tested our approach on the synthetic datasets and compared 
the causal graph performance (AUROC) with three definite causal discovery 
SOTA approaches. Because GIN and LFCM cannot recover 
the relationship between observed nodes, we computed the causal graph 
by subtracting $\mathcal{C}$ from structure $\mathcal{H}$.

In Table~\ref{tabauroc}, ours and DAG-GNN comprehensively perform better 
than other definite causal discovery work in all ingredient settings 
because multi-skeleton data is an adverse condition for 
traditional causal discovery. They have to construct different 
causal structures here for different skeletons, 
which leads to most samples being unavailable in a particular structure. 
Besides, with the decrease in \textbf{Sample} of per skeleton 
($n \in \{50,10,5\}$ in Syn3, Syn2, and Syn1, respectively), 
ours is relatively resilient and steady, but existing approaches 
have a significant decline. These pure independence-based algorithms 
deservedly get stuck in low statistic strength data. Finally,  
ours comprehensively performs better with lower std than DAG-GNN, 
which indicates 
that the estimation of causal strength distribution is exactly a 
surrogate bound to those of independent noise distribution. 

Meanwhile, we find that cluster-based algorithms (GIN and LFCM) 
had a contrary performance when the number of \textbf{Confounders} 
and the \textbf{Pervasiveness} of confounding changed. These results 
corroborate the theory of each algorithm: GIN and LFCM concentrate 
the node cluster induced by different confounders more, while pcss 
more benefit from the pervasive confounding. However, biCDs guarantee 
soundness in these two ingredient settings 
since the confounding effect can unveil 
the true causal structure in a confounding-agnostic solution. 

Additionally, we have shown Figure 2 in Appendix. G.1 
to visualize the causal graphs of 
each approach in the test set. Our method obviously has 
the most approximate estimation of the true causal graph.

\subsubsection{Disentanglement Performance (RQ2)}

We are also concerned about the accuracy of the estimation of graph $\mathcal{C}$ 
to justify our causal disentanglement, 
so we quantify the mean-squared estimation (MSE) error of $C$. 
In Figure~\ref{figmse}, our method likewise performs best in 
all ingredient settings, demonstrating that 
our confounding disentanglement pool the statistical strength 
better than other estimation algorithms in multi skeleton data. 
Besides, combined with the conclusion in
~\cite{agrawal2021decamfounder}, this error should decrease as 
the number of samples $n$ increases. Figure~\ref{figmse} (a) is 
exactly indicative of this conclusion. 
Additionally, to observe the distribution of graph $\mathcal{O}$ 
and graph $\mathcal{C}$, we visualize the 2-D cluster of 
$\widehat{X} $ and $E$ in the varying of all ingredient settings. 
Appendix G.2 shows the projection of $\widehat{X}$ and 
$E$ extracted from the observation relation decoder 
and estimation function, respectively, 
using t-SNE~\cite{knyazev2019understanding}. The results indicate that our 
approach successfully disentangled the confounders and observed nodes 
through the true causal structure. 

\subsection{Real-World Data} 
\begin{wraptable}{r}{0.6\textwidth}
  \centering
  \resizebox{0.6\textwidth}{!}{
  \begin{tabular}{|c|c|c|c|c|c|c|}
    \hline
    Modal&Method&Daily Dialog &MELD&EmoryNLP&IEMOCAP&RECCON\\
    \hline
    \multirow{5}{*}{Text}&Baseline&54.31&60.91&34.42&65.22&63.51\\
    &ACD&55.47&59.47&35.18&65.13&65.29\\
    &CACD&59.53&\textbf{63.81}&39.54&69.17&73.17\\
    &DAG-GNN&57.19&61.48&38.22&67.85&71.36\\
    \cline{2-7}
    &biCD &\textbf{59.68}&63.24&\textbf{40.55}&\textbf{69.86}&\textbf{77.42}\\
    \hline
    \multirow{5}{*}{Video-Audio}&Baseline&$-$&58.30&$-$&56.49&51.87\\
    &ACD&$-$&58.67&$-$&63.27&68.26\\
    &CACD&$-$&60.18&$-$&68.47&70.83\\
    &DAG-GNN&$-$&61.22&$-$&66.83&73.21\\
    \cline{2-7}
    &biCD &$-$&\textbf{64.62}&$-$&\textbf{69.47}&\textbf{76.59}\\
    \hline
  \end{tabular}}
  \caption{Overall performance in Real-world dataset.}
  \label{tabelreal}
\end{wraptable}
To evaluate the quality of the deconfounding representation of $\widehat{X} $ 
that applies in further downstream tasks, we make the assessments 
in the real-world dataset of indefinite data. According to the 
endeavor of~\cite{chen2023affective}, affective reasoning tasks are 
adequate for investigating causal representation. The 
datasets are introduced in Appendix E. 

In Table~\ref{tabelreal}, ours has a much better 
representation ability than other models. Although ACD and CACD 
can recover causal representation from indefinite data, 
ACD only applies to the stationary confounders, and CACD is unable to 
deconfounding. Besides, it is interesting to observe that DAG-GNN lags 
further behind the ours. This performance differs from 
the synthetic definite data, indicating that 
the independent noise estimation is inadequate for indefinite data. 
\begin{wraptable}{r}{0.6\textwidth}
  \centering
  \resizebox{0.6\textwidth}{!}{
  \begin{tabular}{|c|c|c|c|c|c|}
    \hline
    \multirow{2}{*}{Model}&\multicolumn{5}{c}{Datasets}\\
    \cline{2-6}
    &Daily Dialog &MELD&EmoryNLP&IEMOCAP&RECCON\\
    \hline
    Ours&59.68&63.24&40.55&69.86&77.42\\
    \hline
    w/o $\omega(L)$&$\downarrow$1.12&$\downarrow$2.11&$\downarrow$0.85&$\downarrow$3.09&$\downarrow$2.29\\
    w/o $z$&$\downarrow$1.57&$\downarrow$1.58&$\downarrow$1.64&$\downarrow$2.33&$\downarrow$3.74\\
    w/o $C$ estimation&\textbf{$\downarrow$3.84}&\textbf{$\downarrow$2.45}&\textbf{$\downarrow$2.31}&\textbf{$\downarrow$4.61}&\textbf{$\downarrow$4.79}\\
    \hline
  \end{tabular}}
  \caption{Ablation results. w/o $\omega(L)$ means without 
  $\omega(L)=rank(L)/N$ in the reconstruction loss, w/o $z$ means 
  $z=E$ rather than $z =I-A$, and w/o $C$ means 
  without $C$ estimation function.}
  \label{tabablation}
\end{wraptable}

To validate the importance of each module in our method, in Table
~\ref{tabablation}, we conduct ablation studies `w/o $\omega(L)$', 
`w/o $z$', and ` w/o $C$ estimation'. $\omega(L)$ is important for 
weighing confounding samples but removing it causes a performance decrease. 
Furthermore, sampling from adjacent matrix guarantees that our model can learn 
the invariance of multi-skeleton data, so removing $z$ also hampers the 
structure learning. Finally, `w/o $C$' decreases most heavily, 
indicating that confounding disentanglement is crucial for recovering 
the true causal relations. 

\section{Discussion}~\label{AppH}

\begin{wraptable}{r}{0.4\textwidth}
  \centering
  \resizebox{0.4\textwidth}{!}{
  \begin{tabular}{|c|c|c|}
    \hline
    Causal Structure&Attribute&Edge\\
    \hline
    $\mathcal{O}$&known ($\widehat{X}$)&known($A$)\\
    $\mathcal{C}$&substitute ($C$)&unknown\\
    $\mathcal{G}$&partially known&unknown\\
    \hline
  \end{tabular}}
\label{discussiontable}
\caption{Which goals do our model achieve?}
\end{wraptable}

\subsection{achievement of Our Method}

Our method tries to recover entire causal relations including 
observed variables and latent confounders, while this appetite 
for indefinite data has been partially addressed. 

Table~\ref{discussiontable} summarizes the achievement of our method 
in three causal structures. For $\mathcal{O}$, pure causal relations 
between observed variables, our method is capable of gaining attribute  
and edge. Our experiment also demonstrates that it does build a subgraph 
without confounding via $\widehat{X}$ as nodes and $A$ as edges. 
However, our method only gains esoteric knowledge for confounders via 
computing confounding effect $C$. In other words, we do not know 
where and how many confounders are, which leads to the puzzle when we 
try to draw an edge from latent confounders to observed variables. 
Similarly, the unknown of latent confounders prevents us from 
understanding and computing the entire causal structures, thus hampering 
their construction. 

This is also a prevalent obstacle existing in other causal discovery methods. 
Intrinsically, we can not obtain precise information on latent 
variables without any extra knowledge so we have to make many 
assumptions, set many conditions or find substitutes. 
Nevertheless, our endeavor 
of causal discovery in the presence of latent confounders has 
continued to show the feasibility of recovering entire relations from 
purely observational, non-experimental data. 

\begin{figure*}[h]
  \centering
  \subfigure[$\mathcal{H}$]{
    \includegraphics[width=0.22\textwidth]{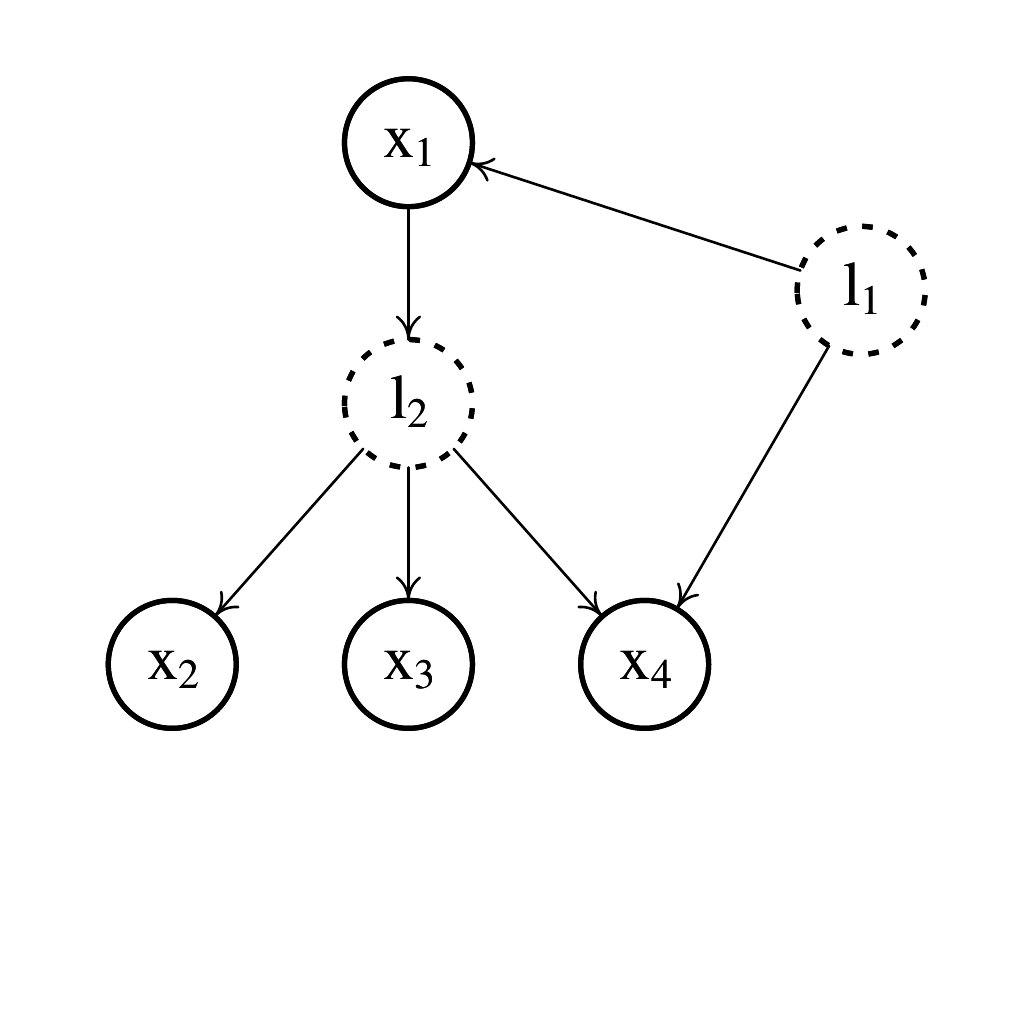}}
  \subfigure[$\mathcal{C}$]{
    \includegraphics[width=0.22\textwidth]{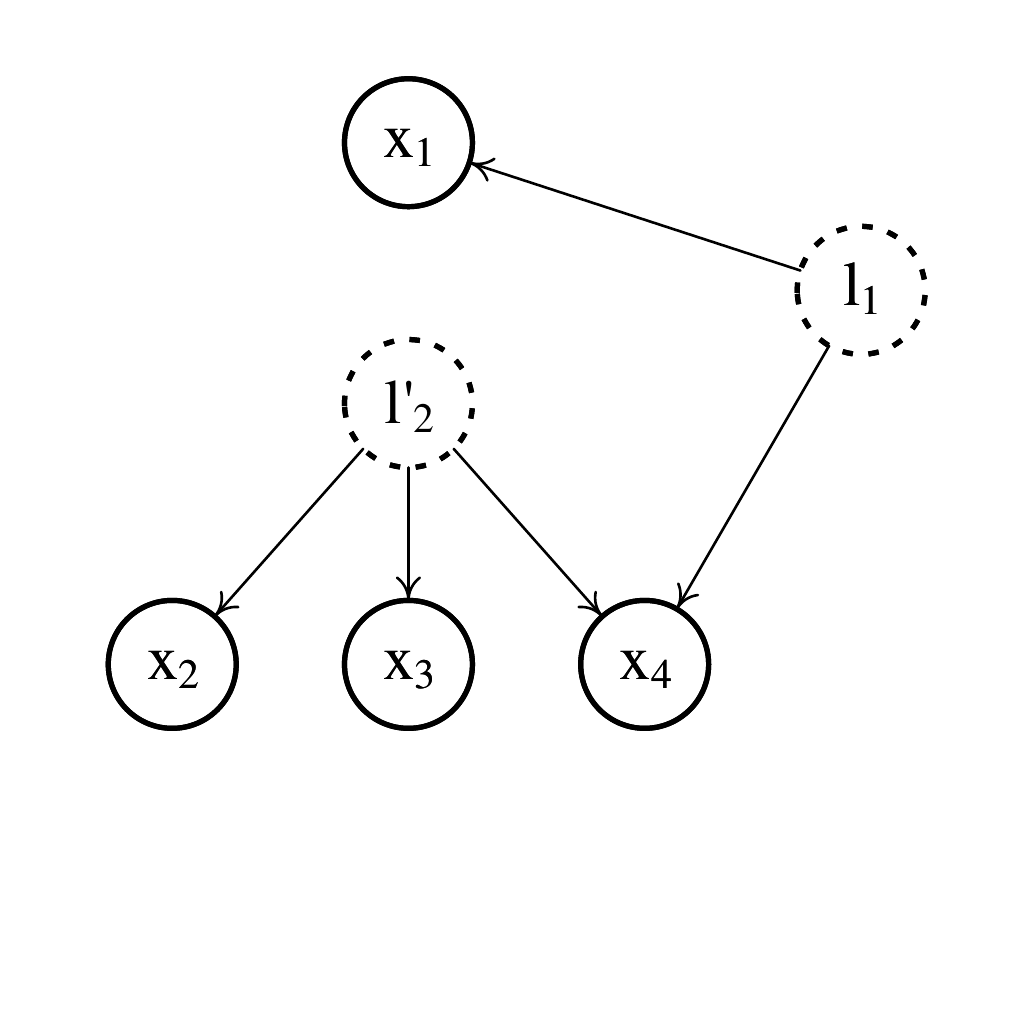}}
  \subfigure[$\mathcal{O}$]{
    \includegraphics[width=0.22\textwidth]{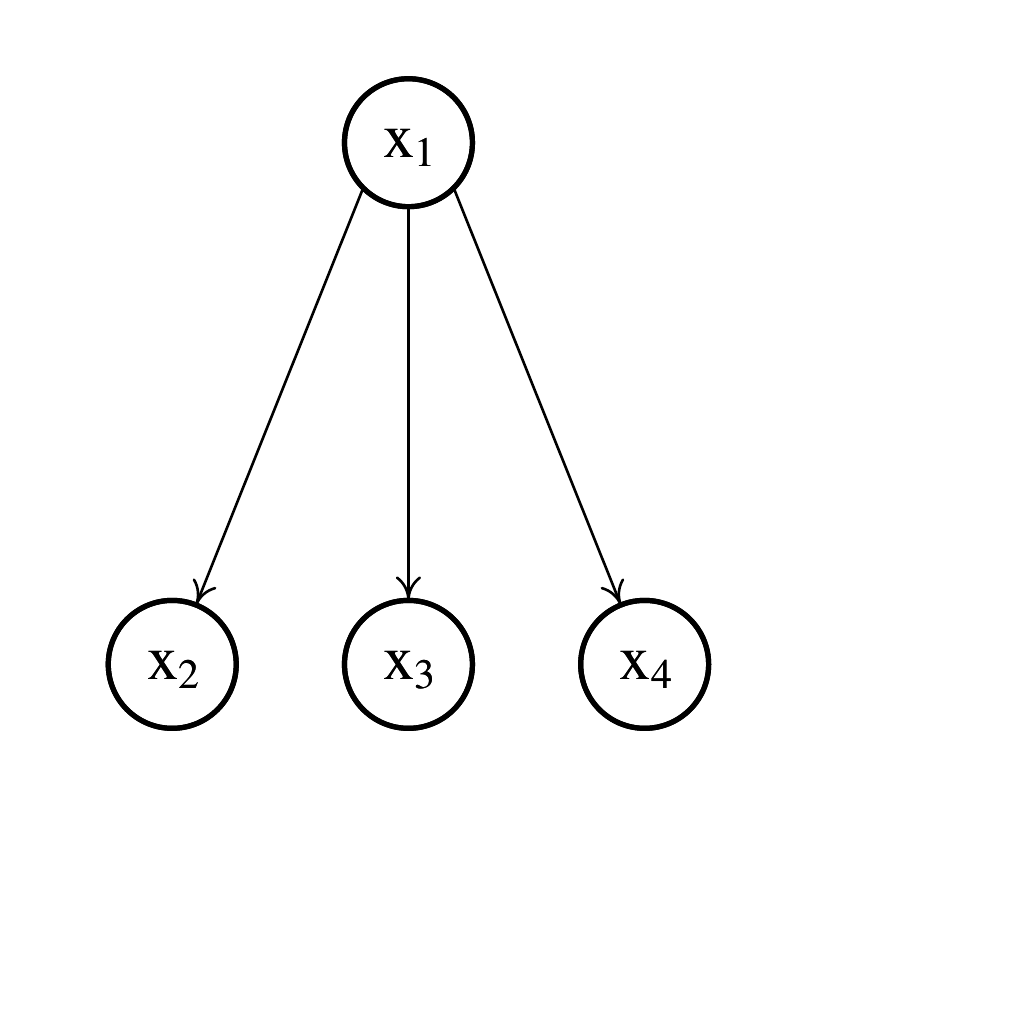}}
  \subfigure[$\mathcal{H\ast }$]{
    \includegraphics[width=0.22\textwidth]{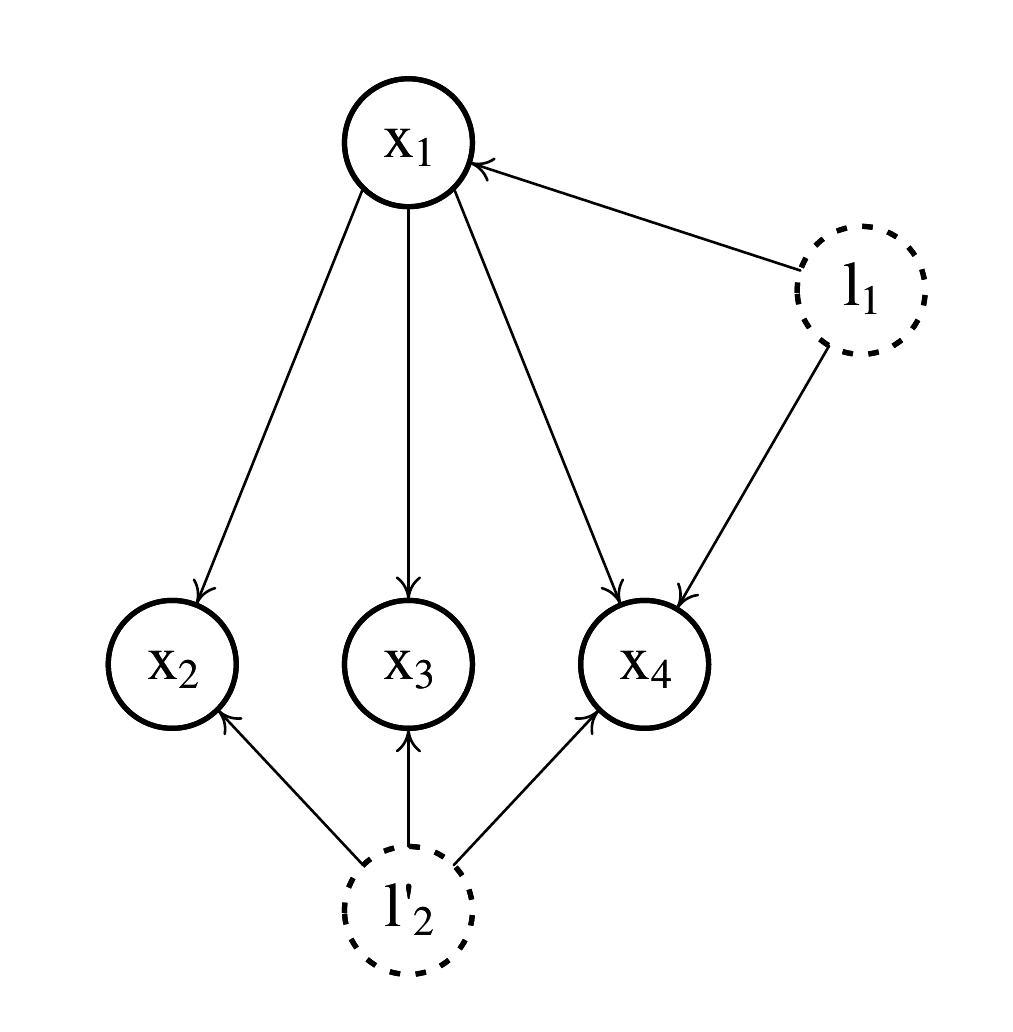}}
  \caption{How our model work when the latent confounders are not 
  exogenous.}
  \label{figprojection}
\end{figure*}

\subsection{Endogenous Confounders}

Besides, another discussion is to explain the situation when 
latent confounders are not exogenous variables. Our model, especially 
two disentangled subgraphs, is based on an underlying assumption: 
each latent variable is exogenous, i.e., there is no edge 
like `$x_{1} \to l_{1}$'. So it is a problem of how our method works 
when there is an edge from an observed variable to 
a latent confounder.

In fact, our method achieve this problem by some projections as shown 
in Figure~\ref{figprojection}. We assume that $l_{1}=\epsilon_{l_1}$, 
$x_{1}=\alpha_{1}l_{1}+\epsilon_{x_{1}}$, 
$l_{2}=\alpha_{2}x_{1}+\epsilon_{l_{2}}$, 
$x_{2}=\alpha_{3}l_{2}+\epsilon_{x_{2}}$ following the SCMs. 
In graph $\mathcal{C}$, we can obtain that: 

\begin{align}
  x_{1}=&\alpha_{1}l_{1}\\
  x_{2}=&\alpha_{3}(\alpha_{2}\alpha_{1}\epsilon_{l_1}+\epsilon_{l_{2}})\\
       =&\alpha_{3}l^{'}_{2}\\
\end{align}

In graph $\mathcal{O}$, we can obtain that: 

\begin{align}
  x_{2}=&\alpha_{3}\alpha_{2}\epsilon_{x_{1}}+\epsilon_{x_{2}}\\
       =&\alpha_{\mathcal{O}}x_{1}+\epsilon_{x_{2}}\\
\end{align}

Overall, there is an underlying projection when our method handles 
these skeletons, which constructs an equal latent confounder $l^{'}$ to 
displace endogenetic $l$ and build direct edges from ancient 
observed nodes in the path when latent confounding is blocking.

\section{Conclusion}

In this paper, we first introduce two key traits-
\textit{skeleton quantity} and \textit{variable dimension}, and 
three data paradigms of Causal Discovery. Then to learn the causal and 
deconfounding representation from indefinite data, 
we design a dynamic varaitional framework, which includes 
a deduction of the available variational posterior $q((I-A)|X)$ and 
analysis of dynamic reconstruction loss. The well-designed variational 
posterior settle down two essential problems induced by indefinite 
data (i.e., multi skeletons and multi values). 
Furthermore, the novel reconstruction describes the 
causal representation dynamic varying with the 
confounding effect's strength.  Moreover, we detailed comprehensive 
experiments on synthetic and real-world data to gauge the effectiveness 
of causal graph identification, causal disentanglement, 
and the high level performance of causal representation. 
Finally, we discuss the limitations about unknown information about 
causal graph and endogenous confounders. 
From the promising results, we thus argue that dynamic variational 
framework is our 
\textit{de facto} need for causal discovery in an 
indefinite data paradigm. 

\bibliographystyle{abbrvnat}
\bibliography{aaai24}

\appendix

\subsection{Structural Causal Model (SCM)}~\label{scm}

The causal additive model (CAM) has some desirable properties, making 
it a suitable method for introducing low-dimensional structures to improve 
statistical efficiency and flexibility
~\cite{buhlmann2014cam,shimizu2014bayesian}. To attempt to unveil the 
causal structure of indefinite data, we will first provide an 
interpretation of SCMs in indefinite data. We begin with a general 
CAM for definite data, assuming that any node $X_{s}$in definite data 
satisfies the following generating process: 
\begin{equation} 
  x_{s,j}=\sum_{x_{s,i}\in Pa(x_{s,j})} f_{s,ij} x_{s,i} +\epsilon_{x_{s,j}} (x \in \mathbb{R}^{N \times 1},\epsilon_{x_{s,j}} \in \mathbb{R}^{N \times 1}, 0\leq i \neq j < N)
  \label{eqt1}
 \end{equation}
 where $\epsilon_{x_{s,j}}$ represents the independent and identically 
 distributed noise variable (Existing methods support it satisfies 
 some particular distribution assumption, including both non-gaussian 
 distribution~\cite{shimizu2006linear} or gaussian distribution
 ~\cite{agrawal2021decamfounder}). $f_{s,ij}$ represents the causal 
 strength from node $x_{s,i}$ to $x_{s,j}$ in $s$-th sample. 
 $Pa(x_{s,j})$ is the parent set of $x_{s,j}$. Without loss of 
 generality, we can omit the sample number $s$. Continuing equation
 ~\ref{eqt1}, we make CAM assumption for indefinite data and consider 
 the following generating process:
 \begin{equation} 
  x_{m,j}=\sum_{x_{m,i}\in Pa(x_{m,j})} f_{m,ij} x_{m,i} +\epsilon_{x_{m,j}} (x \in \mathbb{R}^{N \times D},\epsilon_{x_{s,j}} \in \mathbb{R}^{N \times D},, 0\leq i \neq j < N)
  \label{eqt2}
 \end{equation}
 where $\epsilon_{x_{j}}$ is a distribution-agnostic noise variable. 
 $x_{m,i}$ represents the $i$-th multi-value node in $m$-th skeleton. 
 Consequently, we can obtain the general definition of SCM between any 
 sample in definite data and any sample in one skeleton of indefinite data: 
 \begin{definition}[Structural Causal Model]
  An SCM is a 3-tuple $\langle X_{m},\mathcal{F}_{m},\mathbb{P}_{m}\rangle $, where 
  $X_{m}$ is the entire set of observed variables 
  $X_{m}=\{x_{m,i}\}^{S}_{i=1}$. Structural equations 
  $\mathcal{F}_{m}=\{f_{m,i}\}^{S}_{i=1}$ are functions that determine 
  $X$ with $x_{m,i}=f_{m,i}(Pa(x_{{m,i}}))$, 
  where $Pa(x_{{m,i}})\subseteq X_{m}$. 
  $\mathbb{P}_{m}(X)$ is a distribution over $X_{m}$. 
  \label{def3}
\end{definition}

\section{Proof of Definition 3} \label{AppD}
By equation 7 in main text and~\ref{eqt2} in Appendix, 

\begin{equation}
  (x_{j}-C_{j})=\sum_{x_{i}\in Pa(x_{j})} A_{i,j} (x_{i}-C_{i})+\epsilon_{j}
\end{equation}

Hence, 

\begin{align}
  C_{j}=&\sum_{x_{i}\in Pa(x_{j})} A_{i,j}C_{i}+B_{j}L\\
       =&\sum_{x_{i}\in Pa(x_{j})} A_{i,j}[(I-A)^{-1}BL]_{i}+B_{j}L
       \label{eqt41}
\end{align}

Intuitively, the term $(I-A)^{-1}BL$ reflects a particular graph 
$\mathcal{Q}$ consisting of $p(X_{s}|L)$. Hence, we would like to 
transform Equation~\ref{eqt41} into a statistic about `$p(X|L)$'. 
Fortunately, a sufficient statistic $C_{j}=E(X_{j}|L)$ is supported 
by~\cite{agrawal2021decamfounder}. It makes $\mathcal{Q}$ identifiable 
under the Gaussian $\epsilon$ and $E(\epsilon)=0$. Along with their 
inference, we find the branch point under our data condition: 

\begin{align}
  C_{j}=&\sum_{x_{i}\in Pa(x_{j})} A_{i,j}[(I-A)^{-1}BL]_{i}+B_{j}L\\
      =&E[\sum_{x_{i}\in Pa(x_{j})} f_{ij}[(I-A)^{-1}E+(I-A)^{-1}BL]_{i}+B_{j}L|L]\\
\end{align}

If the $E(\epsilon)$ is negligible, we can obtain the same statistic 
due to $C_{j}=E[\sum_{x_{i}\in Pa(x_{j})} f_{ij}[(I-A)^{-1}E+(I-A)^{-1}BL]_{i}+B_{j}L+\epsilon_{j}|L]$. 
In other words, when $E(L)$ apparently exceeds $E(\epsilon)$, latent 
confounders essentially contribute the $C$, and naturally, $C$ is 
meaningless when it is mainly affected by $\epsilon$. 

Finally, we could approximately estimate the discrete probability 
of $C$ under a strong confounding assumption: 

\begin{align}
  C_{j}=&E(x_{j}|L)\\
       =&\frac{\mathbb{P}(x_{j}) \mathbb{P}(L|x_{j})}{\sum_{i}^{N} \mathbb{P}(x_{i}) \mathbb{P}(L|x_{i}) } x_{j} 
\end{align}

Additionally, $C$ is not easy to calculate under weak confounding, 
so the dynamic reconstruction loss function does not consider C 
when confounding influence is insufficient 
(See Subsection ``Dynamic Reconstruction Error'' for details).

\section{Datasets}\label{AppE}

Synthetic datasets afford several advantages not present in 
real-world datasets. For instance, causal relationships 
within synthetic datasets remain unbiased; synthetic datasets 
grant control over the sample size of each structure 
within the multi-skeleton setting. 
In addition, complete labels for latent confounding 
can be included in synthetic datasets; 
along with the capacity to dictate variations in other parameters, 
such as the quantity of observed nodes and confounders. 
However, synthetic datasets also exhibit certain limitations, 
primarily their inability to construct multi-value variables. 
This arises due to the complex interdependencies 
between values within these variables, which cannot be 
defined through distribution assumptions. 

Therefore, to conduct the proposed experiment, 
we adopt an approach incorporating both synthetic and 
real-world datasets. The synthetic datasets function as 
semi-definite data involving multi-skeleton and single-value data. 
This setup seeks to evaluate the causal discovery capabilities 
of our proposed methods and their ability to decouple 
confoundings under various conditions under multi skeletons. 
The real-world datasets, composed of indefinite data 
with multi-skeleotn and multi-value, aim to measure 
the actual enhancement provided by our model-generated 
causal representations for high-level tasks. 
Table~\ref{tabdataset} shows the detailed statistic about synthetic 
datasets and real-word datasets.

\subsection{Synthetic Data}

\begin{table}
  \footnotesize
  \centering
  \resizebox{\linewidth}{!}{
    \begin{tabular}{|c|c|c|c|c|c|c|c|}
      \hline
    \multirow{2}{*}{Category}&\multirow{2}{*}{Dataset}&\multicolumn{2}{c}{Train Set}&\multicolumn{2}{c}{Valid Set}&\multicolumn{2}{c}{Test Set}\\
    \cline{3-8}
    &&$\#$sample&$\#$skeleton&$\#$sample&$\#$skeleton&$\#$sample&$\#$skeleton\\
    \hline
    \multirow{4}{*}{Synthetic}&Syn1&2250&450&500&100&1000&200\\
    &Syn2&4500&450&1000&100&2000&200\\
    &Syn3&22500&450&5000&100&10000&200\\
    &Syn4-9&22500&450&5000&100&10000&200\\
    \hline
    \multirow{6}{*}{Real-world}&DailyDialog&11118&35&1000&25&1000&23\\
    &MELD&1038&14&114&7&280&9\\
    &EmoryNLP&713&31&99&14&85&11\\
    &IEMOCAP&100&13&20&4&31&5\\
    &RECCON-DD&833&37&47&15&225&28\\
    &RECCON-IE&$-$ &$-$&$-$ &$-$&16&14\\
    \hline

  \end{tabular}}
  \caption{Statistics of Synthetic and Real-world datasets}
  \label{tabdataset}
\end{table}

\begin{table}
  \footnotesize
  \centering
  \resizebox{0.7\linewidth}{!}{
  \begin{tabular}{|c|c|c|c|c|}
    \hline
    \textbf{ID}&\textbf{$\#$sample} &\textbf{$\#$confounder}&\textbf{$\#$observed nodes}&\textbf{pervasiveness}\\
    \hline
    Syn1&5&1&50&0.7\\
    Syn2&10&1&50&0.7\\
    Syn3&50&1&50&0.7\\
    Syn4&50&5&50&0.7\\
    Syn5&50&10&50&0.7\\
    Syn6&50&1&20&0.7\\
    Syn7&50&1&100&0.7\\
    Syn8&50&1&50&0.1\\
    Syn9&50&1&50&0.4\\
    \hline
  \end{tabular}}
  \caption{The key ingredients of each synthetic dataset}
  \label{tabelingredient}
\end{table}

 Specifically, We randomly draw Causal DAG 
from a random graph model with an expected neighborhood size of 5 and 
consider graphs with the number of observed nodes $N\in \{20, 50, 100\}$. 
For probing how our approach is affected by the pervasiveness of 
confounding, we assume that each confounder $l_{k}$ is a direct cause 
of node $x_{i}$ with a chance $P \in \{0.1, 0.4, 0.7\}$. Given the graph, 
we stochastically set a trend type for each causal strength weight 
with equal probability. Meanwhile, we add $N(0,\sigma^{2}_{noise})$ 
noise to each node. Finally, we consider the number of confounders 
$K \in \{1, 5, 10\}$ and the number of samples of each skeleton 
$n \in \{5, 10, 50\}$, respectively. These key ingredients of each set of 
synthetic data are shown in Table~\ref{tabelingredient}.

\subsection{Real-World Data} 

We use five real datasets for two affective reasoning tasks. These 
datasets are conducted by Conversational Affective Causal Discovery 
(CACD)~\cite{chen2023affective}. 

\textbf{DailyDialog}~\cite{li-etal-2017-dailydialog}:
A Human-written dialogs dataset with 7 emotion labels (\textit{neutral}, 
\textit{happiness}, \textit{surprise}, \textit{sadness}, 
\textit{anger}, \textit{disgust}, and \textit{fear}). We follow 
~\cite{shen-etal-2021-directed} to regard utterance turns as the speaker turns. 

\textbf{MELD}~\cite{poria-etal-2019-meld}: 
A multimodel ERC dataset with 7 emotion labels as the same as DailyDialog. 

\textbf{EmoryNLP}~\cite{zahiri:18a}:
A TV show scripts dataset with 7 emotion labels (\textit{neutral}, 
\textit{sad}, \textit{mad}, \textit{scared}, \textit{powerful}, 
\textit{peaceful}, \textit{joyful}). 

\textbf{IEMOCAP}~\cite{busso2008iemocap}: 
A multimodel ERC dataset with 9 emotion labels (\textit{neutral}, 
\textit{happy}, \textit{sad}, \textit{angry}, \textit{frustrated}, 
\textit{excited}, \textit{surprised}, \textit{disappointed}, and \textit{fear}). 
However, models in ERC field are often evaluated on samples with 
the first six emotions due to the too few samples of the latter three emotions. 
20 dialogues for validation set is following~\cite{shen-etal-2021-directed}. 

\textbf{RECCON}~\cite{poria2021recognizing}: 
The first dataset for emotion cause recognition of conversation 
including RECCON-DD and RECCON-IE (emulating an out-of-distribution 
generalization test). RECCON-DD includes 5380 labeled ECPs and 5 cause 
spans (\textit{no-context}, \textit{inter-personal}, \textit{self-contagion}, 
\textit{hybrid}, and \textit{latent}). 

\section{Baselines and Experiment setup}\label{AppF}

To the best of our knowledge, no baseline approach is 
broadly applicable in single-value and multi-value data, 
even single-skeleton and multi-skeleton datasets. So we choose 
the SOTA work in Causal Discovery from definite and indefinite 
data, respectively. In synthetic data, we investigated \textbf{pcss}
~\cite{agrawal2021decamfounder}, \textbf{LFCM}~\cite{squires2022causal}, 
and \textbf{GIN}~\cite{xie2020generalized}, which are all SOTA approaches in 
Causal Discovery, especially from single-value and single-skeleton 
data. 

\textbf{pcss}: estimates causal relationships in the non-linear, 
pervasive confounding setting by leveraging an approximate spectral 
decomposition of the observed data matrix. 

\textbf{LFCM}: finds ordered clusters of observed nodes, 
a partial ordering over clusters, and finally the entire structure 
over both observed and latent nodes via the theorem of 
Trek separation and Tetrad representation. 

\textbf{GIN}: designs a Linear Non-Gaussian Latent variable Model 
to identify important information, including where the latent variables 
are, the number of latent variables behind any two observed variables, 
and the causal order of the latent variables. 

In Real-world data, we investigated \textbf{DAG-GNN}~\cite{yu2019dag}, 
\textbf{ACD}~\cite{lowe2022amortized} and \textbf{CACD}~\cite{chen2023affective}. 
DAG-GNN is the SOTA method for single-skeleton and multi-value data, 
ACD is the SOTA method for multi-skeleton data while single-value. 
CACD is the SOTA work on multi-skeleton and 
multi-value data while neglecting the confounding. 

\textbf{ACD}: leverages shared dynamics to learn to infer causal 
relationships from multi-skeleton time-series data via a 
single, amortized model. 

\textbf{CACD}: discover causal relationships in multi-value data 
via designing a common skeleton and generating a substitute for 
independent noise.

\textbf{DAG-GNN}: leverages SEM to construct a gnn-based variational 
model adopting independent noise $E$ as latent viariable.  

Due to dissatisfing data structure, ACD and CACD are not capable for 
synthetic data. In synthetic data, we only add DAG-GNN to represent 
indefinite data method. Besides, pcss, LFCM, and GIN are all unavailable 
for real-world dataset. 

\section{Visualization}
\subsection{Causal Strength}\label{AppG1}

\begin{figure}
  \centering
  \subfigure[GIN]{
    \includegraphics[width=0.45\textwidth]{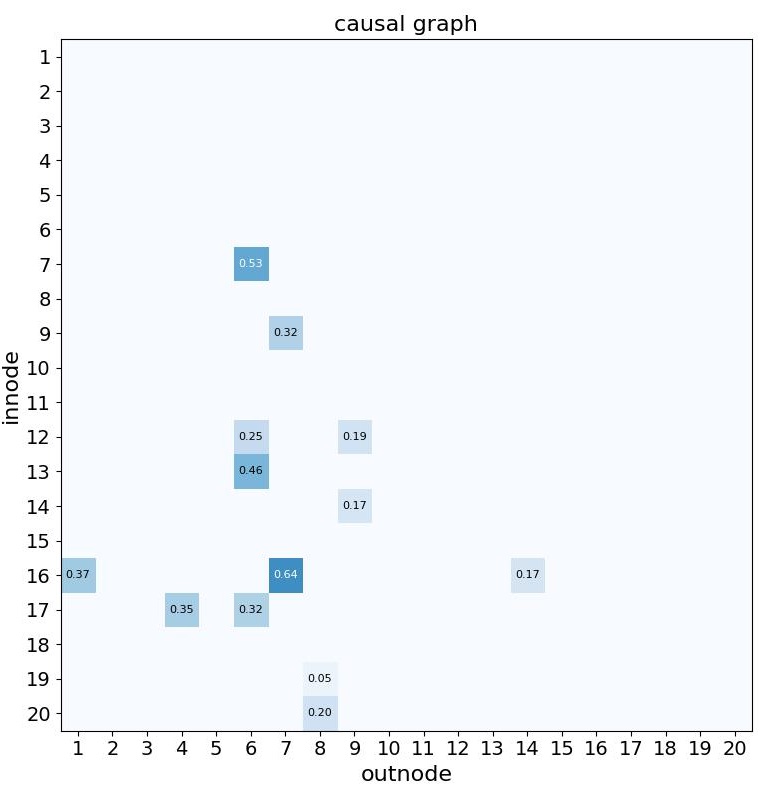}}
  \subfigure[LFCM]{
    \includegraphics[width=0.45\textwidth]{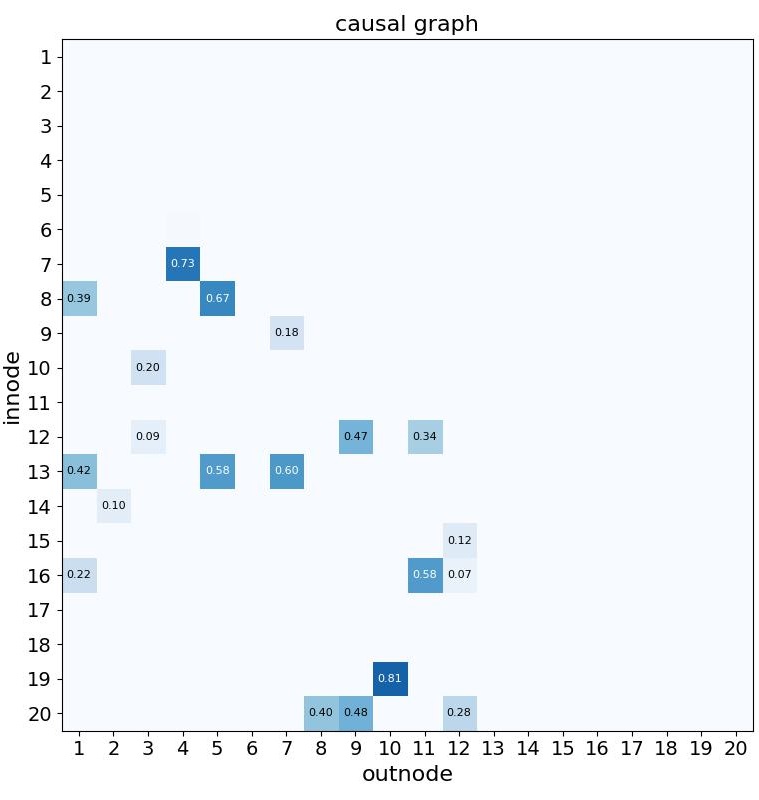}}
  \subfigure[pcss]{
    \includegraphics[width=0.45\textwidth]{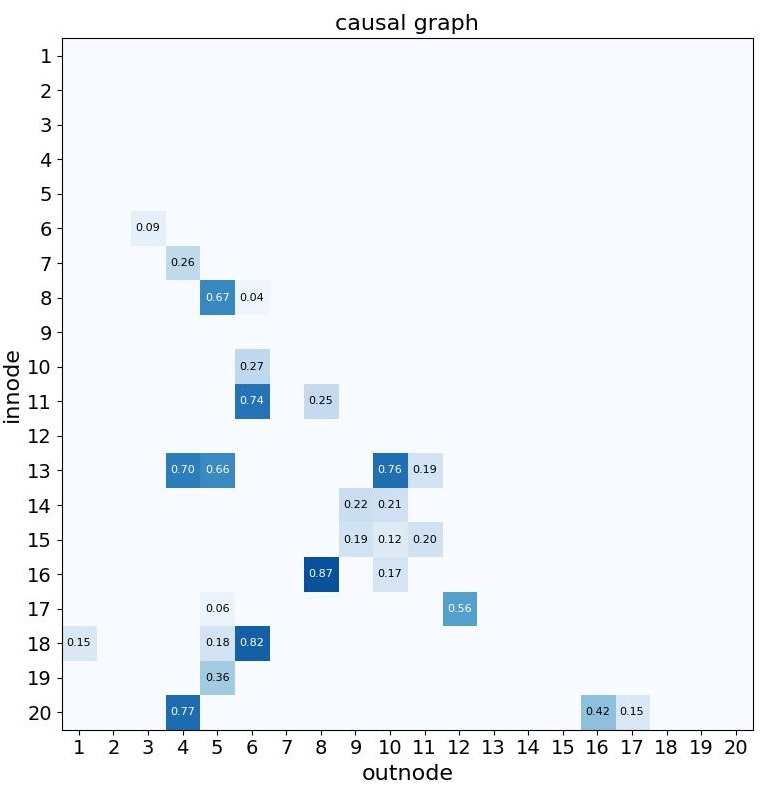}}
  \subfigure[DAG-GNN]{
    \includegraphics[width=0.45\textwidth]{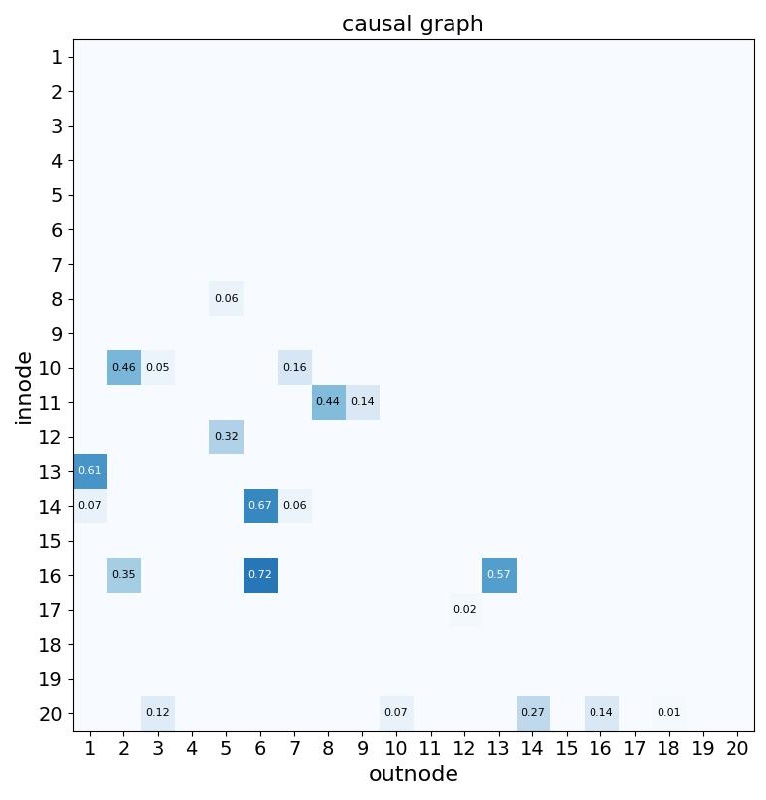}}
  \subfigure[ours]{
    \includegraphics[width=0.45\textwidth]{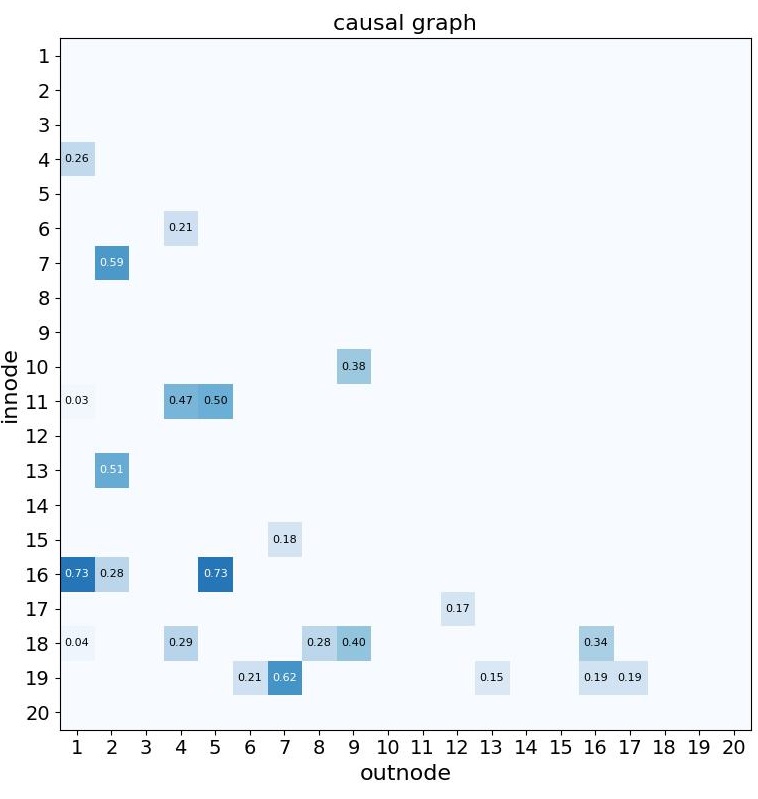}}
  \subfigure[true graph]{
    \includegraphics[width=0.45\textwidth]{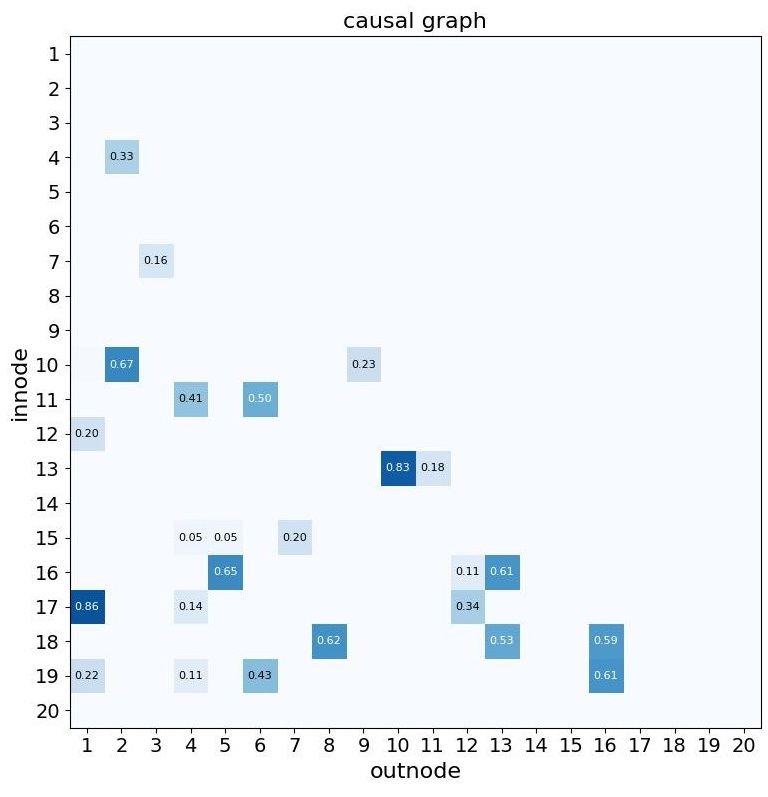}}
  \caption{Visualization of predicted causal graphs (figure a-e) 
  and true graph (figure f) of the test set in Syn3. We use adjacent 
  matrix to represent causal strength. }
  \label{figvis}
\end{figure}

In Figure~\ref{figvis}, we illustrate the causal graph generated 
by various methods, based on samples of 20 observed variables, 
in the form of adjacency matrices. For each  $(i, j)$ in the matrix, 
the value signifies the probability of a causal relationship 
from node $j$ to $i$ ($P \in [0,1]$); therefore, 
the graph is a strict lower-triangular matrix. 
Subfigure f represents the actual causal model, w
hile subfigure a-e display the predictions made by various methods 
concerning this causal structure. Significant similarities 
are observed between our approach (subfigure e) and ground truth f, 
as evidenced by the highest number of correct predictions 
(and the least number of incorrect predictions) 
attained via our method: On average, for a causal structure 
with 20 observational nodes, our method exceeds other approaches 
by predicting over 2 directed edges correctly 
and lessens the error by over 1 predicted directed edge. 
However, this is an unsuitable metric, as it does not consider 
the continuous values of the causal relationships. 
Consequently, we utilize the AUROC for the quantitative experimentation.

\subsection{2-D Cluster Visualization}\label{AppG2}

\begin{figure}
  \centering
  \subfigure[nodes=20]{
    \includegraphics[width=0.32\linewidth]{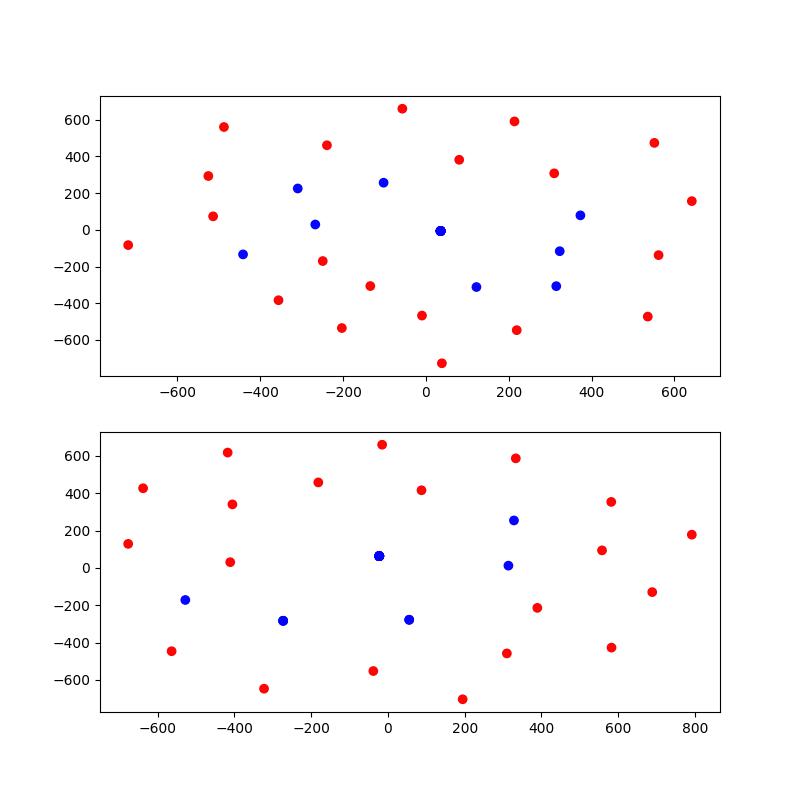}}
  \subfigure[nodes=50]{
    \includegraphics[width=0.32\linewidth]{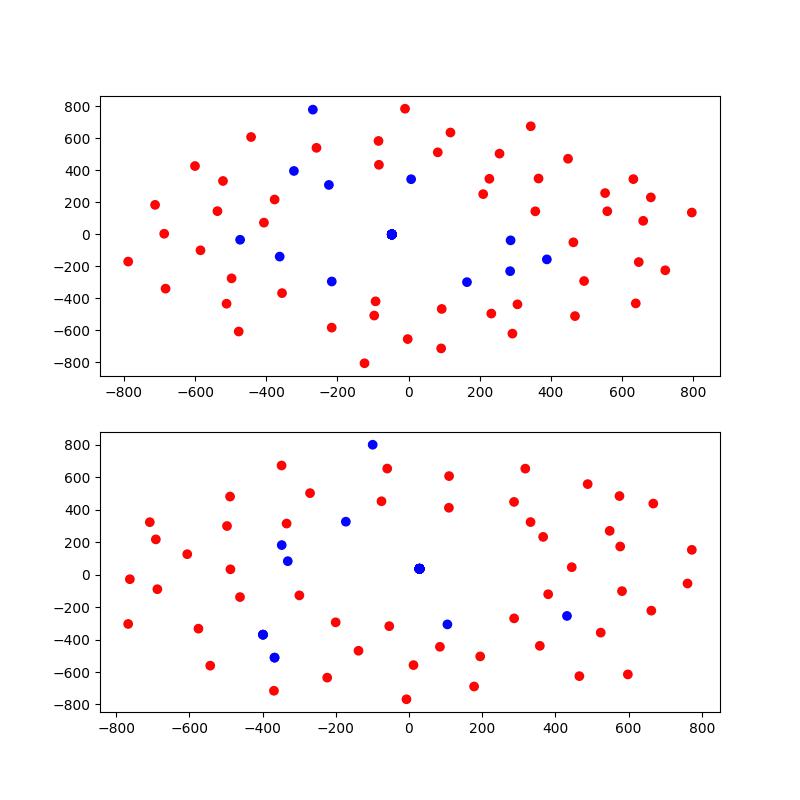}}
  \subfigure[nodes=100]{
    \includegraphics[width=0.32\linewidth]{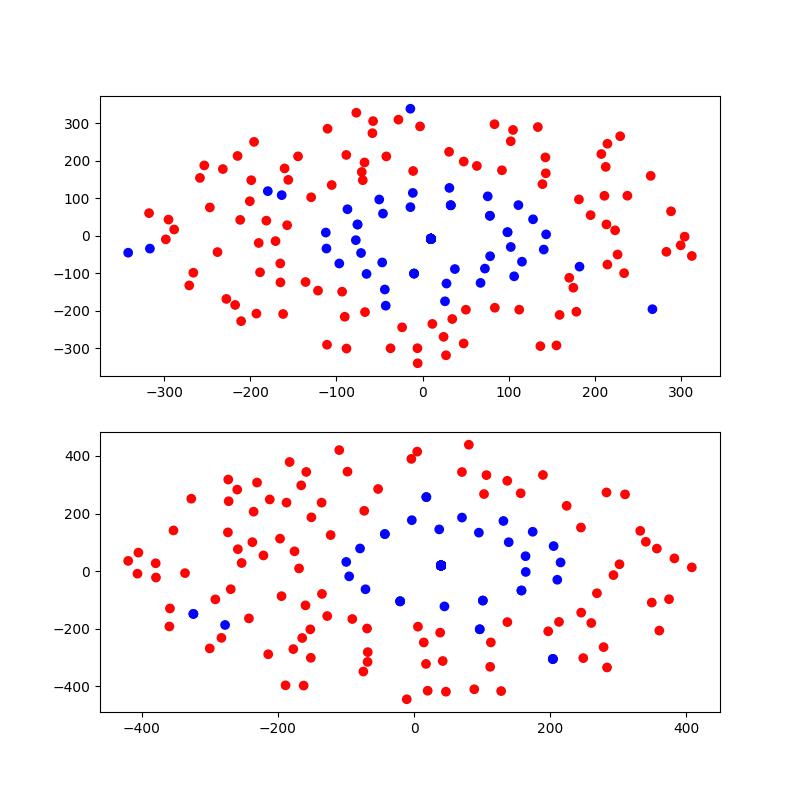}}
  \caption{Visualization of $\widehat{X} $ (red nodes) and 
  $C$ (blue nodes). The upper subfigure represents the 
  predicted results of our approach, and the lower subfigure is the 
  true distribution.}
  \label{figtsne}
\end{figure}
\begin{figure}
  \centering
  \subfigure[samples=5]{
    \includegraphics[width=0.32\linewidth]{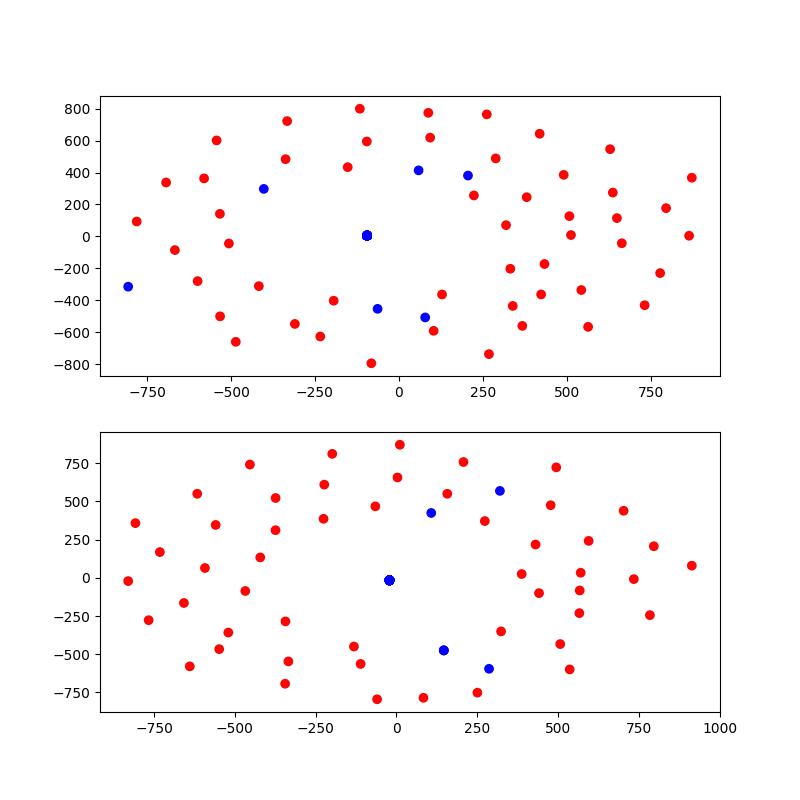}}
  \subfigure[samples=10]{
    \includegraphics[width=0.32\linewidth]{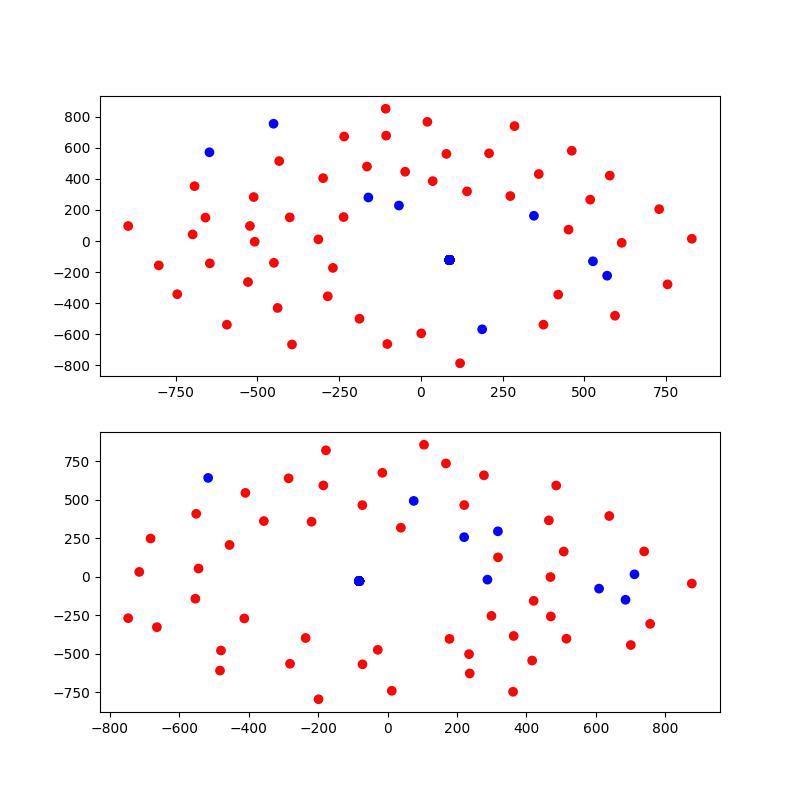}}
  \subfigure[samples=50]{
    \includegraphics[width=0.32\linewidth]{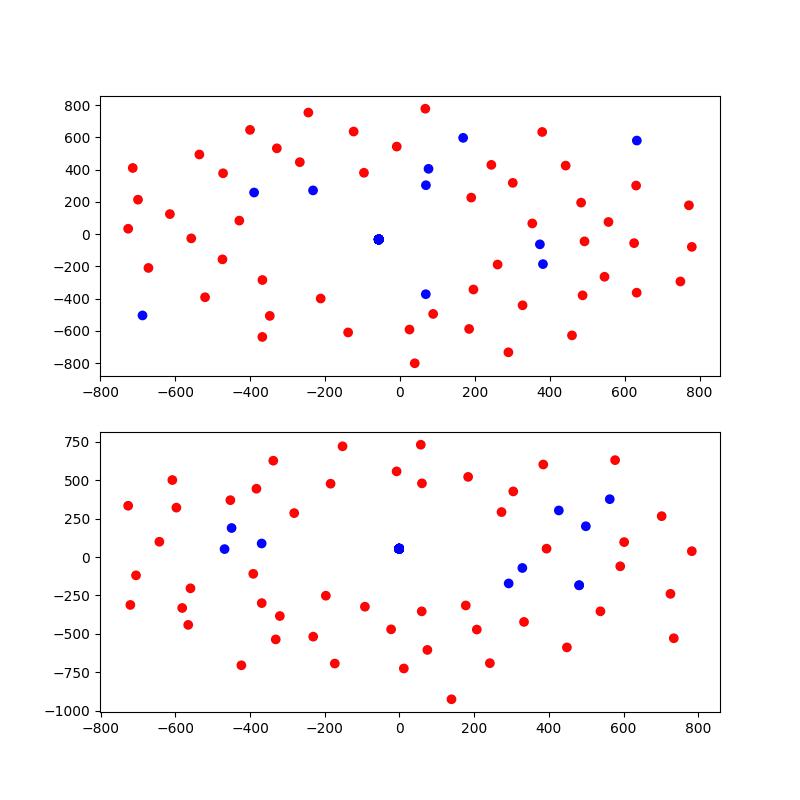}}
  \caption{Visualization of $\widehat{X}$ (red nodes) and 
  $C$ (blue nodes) on Syn1 (a), Syn2 (b), and Syn3 (c).}
  \label{figtsnesample}
\end{figure}

\begin{figure}
  \centering
  \subfigure[confounder=1]{
    \includegraphics[width=0.32\linewidth]{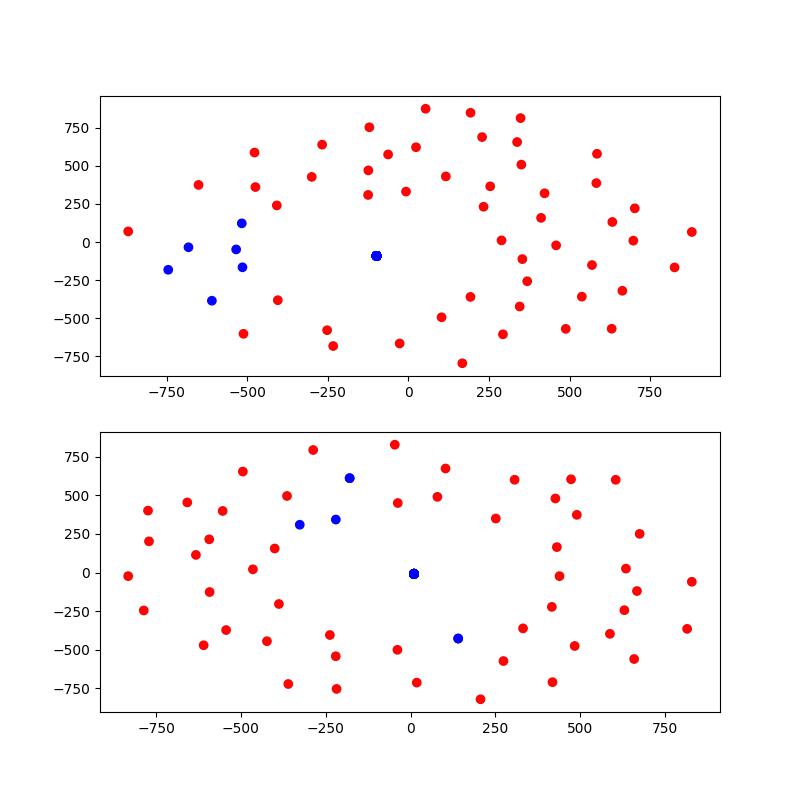}}
  \subfigure[confounders=2]{
    \includegraphics[width=0.32\linewidth]{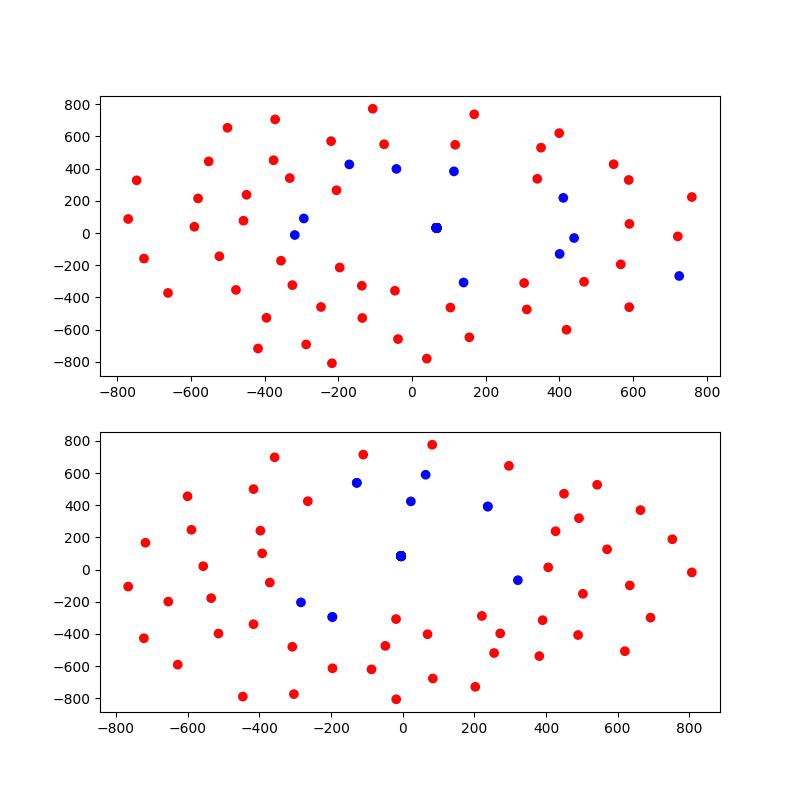}}
  \subfigure[confounders=5]{
    \includegraphics[width=0.32\linewidth]{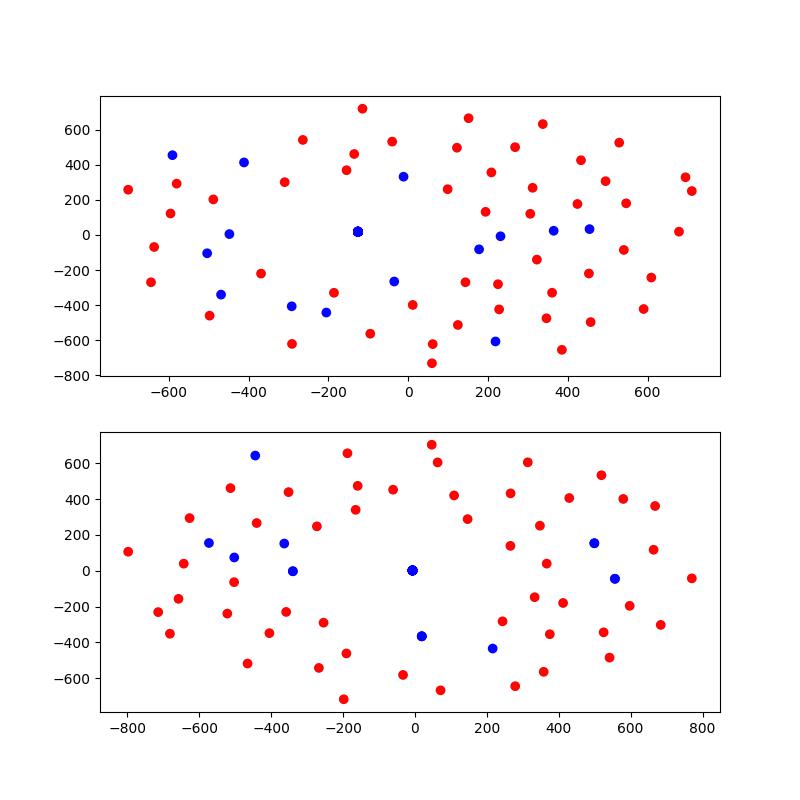}}
  \caption{Visualization of $\widehat{X}$ (red nodes) and 
  $C$ (blue nodes) on Syn3 (a), Syn4 (b), and Syn5 (c).}
  \label{figtsneconfounder}
\end{figure}
\newpage

\begin{figure}
  \centering
  \subfigure[pervasive 0.1]{
    \includegraphics[width=0.32\linewidth]{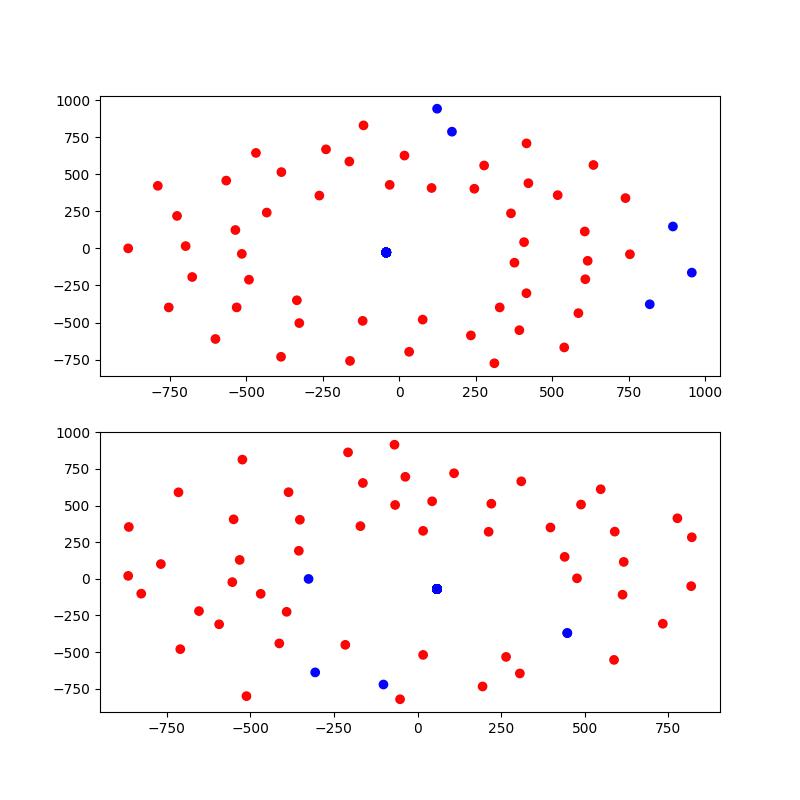}}
  \subfigure[pervasive 0.4]{
    \includegraphics[width=0.32\linewidth]{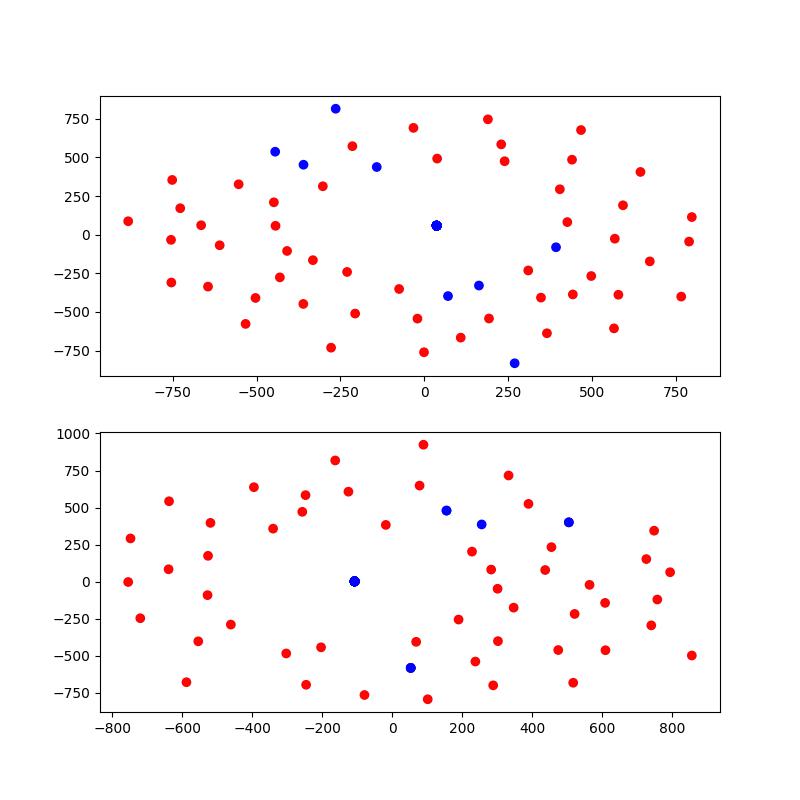}}
  \subfigure[pervasive 0.7]{
    \includegraphics[width=0.32\linewidth]{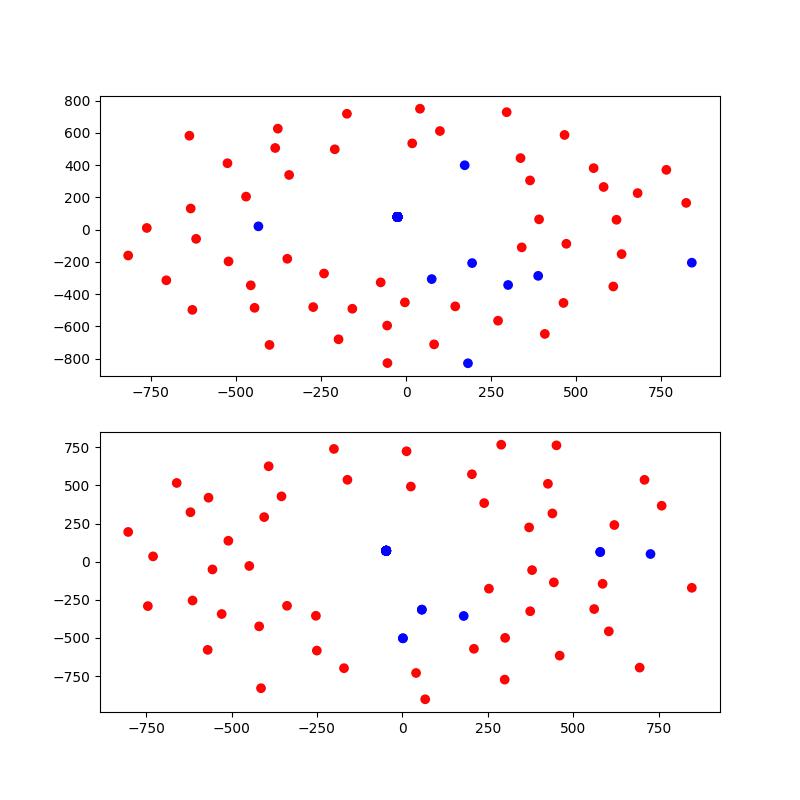}}
  \caption{Visualization of $\widehat{X}$ (red nodes) and 
  $C$ (blue nodes) on Syn3 (a), Syn4 (b), and Syn5 (c).}
  \label{figtsnepervasiveness}
\end{figure}

\end{document}